\documentclass{article}

\usepackage{array}
\usepackage{microtype}
\usepackage{graphicx}
\usepackage{subfigure}
\usepackage{amsmath, bm}
\usepackage{dsfont}
\usepackage{amssymb}
\usepackage{multirow}
\usepackage{booktabs} 
\usepackage{natbib}
\usepackage[vlined,linesnumbered,ruled,algo2e]{algorithm2e}
\usepackage[linesnumbered,ruled,vlined]{algorithm2e}

\SetCommentSty{mycommfont}

\usepackage{varwidth}
\usepackage{pifont}
\usepackage{makecell}

\usepackage{color}
\usepackage[table]{xcolor}
\definecolor{visda_color_0}{rgb}{0.9922,0.0,0.0235}
\definecolor{visda_color_1}{rgb}{0.9922,0.4235,0.0314}
\definecolor{visda_color_2}{rgb}{1.0,1.0,0.3098}
\definecolor{visda_color_3}{rgb}{0.4471,1.0,0.0275}
\definecolor{visda_color_4}{rgb}{0.1686,1.0,0.0980}
\definecolor{visda_color_5}{rgb}{0.1333,1.0,0.4275}
\definecolor{visda_color_6}{rgb}{0.1333,1.0,1.0}
\definecolor{visda_color_7}{rgb}{0.0510,0.4,1.0}
\definecolor{visda_color_8}{rgb}{0.0,0.0,1.0000}
\definecolor{visda_color_9}{rgb}{0.4196,0.0,1.0}
\definecolor{visda_color_10}{rgb}{0.9882,0.0,1.0}
\definecolor{visda_color_11}{rgb}{0.9922,0.0,0.4275}
\definecolor{city_color_0}{rgb}{0.0,0.0,0.0}
\definecolor{city_color_1}{rgb}{0.5020,0.2510,0.5020}
\definecolor{city_color_2}{rgb}{0.9569,0.1373,0.9098}
\definecolor{city_color_3}{rgb}{0.2745,0.2745,0.2745}
\definecolor{city_color_4}{rgb}{0.4000,0.4000,0.6118}
\definecolor{city_color_5}{rgb}{0.7451,0.6000,0.6000}
\definecolor{city_color_6}{rgb}{0.6000,0.6000,0.6000}
\definecolor{city_color_7}{rgb}{0.9804,0.6667,0.1176}
\definecolor{city_color_8}{rgb}{0.8627,0.8627,0.0000}
\definecolor{city_color_9}{rgb}{0.4196,0.5569,0.1373}
\definecolor{city_color_10}{rgb}{0.5961,0.9843,0.5961}
\definecolor{city_color_11}{rgb}{0.2745,0.5098,0.7059}
\definecolor{city_color_12}{rgb}{0.8627,0.0784,0.2353}
\definecolor{city_color_13}{rgb}{1.0000,0.0000,0.0000}
\definecolor{city_color_14}{rgb}{0.0000,0.0000,0.5569}
\definecolor{city_color_15}{rgb}{0.0000,0.0000,0.2745}
\definecolor{city_color_16}{rgb}{0.0000,0.2353,0.3922}
\definecolor{city_color_17}{rgb}{0.0000,0.3137,0.3922}
\definecolor{city_color_18}{rgb}{0.0000,0.0000,0.9020}
\definecolor{city_color_19}{rgb}{0.4667,0.0431,0.1255}
\definecolor{citecolor}{RGB}{119,185,0}

\definecolor{ForestGreen}{rgb}{0.13, 0.55, 0.13}
\definecolor{Maroon}{rgb}{0.69, 0.19, 0.0}
\newcommand{\cmark}{\ding{51}}%
\newcommand{\xmark}{\ding{55}}%

\newcommand{\singlecmark}{\textcolor{Maroon}{\cmark}}
\newcommand{\singlexmark}{\textcolor{ForestGreen}{\xmark}}

\DeclareMathOperator*{\argmax}{arg\,max}
\DeclareMathOperator*{\argmin}{arg\,min}
\newcommand\ddfrac[2]{\frac{\displaystyle #1}{\displaystyle #2}}

\usepackage{hyperref}

\usepackage[accepted]{icml2020}

\icmltitlerunning{Automated Synthetic-to-Real Generalization}

\begin{document}

\twocolumn[
\icmltitle{Automated Synthetic-to-Real Generalization}



\icmlsetsymbol{equal}{*}

\begin{icmlauthorlist}
\icmlauthor{Wuyang Chen}{tamu}
\icmlauthor{Zhiding Yu}{nvidia}
\icmlauthor{Zhangyang Wang}{tamu}
\icmlauthor{Anima Anandkumar}{nvidia,caltech}
\end{icmlauthorlist}

\icmlaffiliation{tamu}{Texas A\&M University}
\icmlaffiliation{nvidia}{NVIDIA}
\icmlaffiliation{caltech}{California Institute of Tech}

\icmlcorrespondingauthor{Wuyang Chen}{wuyang.chen@tamu.edu}
\icmlcorrespondingauthor{Zhiding Yu}{zhidingy@nvidia.com}

\icmlkeywords{Domain Generalization, Domain Adaptation, Synthetic Training, Lifelong Learning, Learning to Optimize}

\vskip 0.3in
]



\printAffiliationsAndNotice{}  

\begin{abstract}
Models trained on synthetic images often face degraded generalization to real data. As a convention, these models are often initialized with ImageNet pre-trained representation. Yet the role of ImageNet knowledge is seldom discussed despite common practices that leverage this knowledge to maintain the generalization ability. An example is the careful hand-tuning of early stopping and layer-wise learning rates, which is shown to improve synthetic-to-real generalization but is also laborious and heuristic. In this work, we explicitly encourage the synthetically trained model to maintain similar representations with the ImageNet pre-trained model, and propose a \textit{learning-to-optimize (L2O)} strategy to automate the selection of layer-wise learning rates. We demonstrate that the proposed framework can significantly improve the synthetic-to-real generalization performance without seeing and training on real data, while also benefiting downstream tasks such as domain adaptation. Code is available at: \url{https://github.com/NVlabs/ASG}.
\end{abstract}

\section{Introduction} \label{sec:intro}
Training a deep convolutional neural network (DCNN) can require large amounts of labeled data in computer vision tasks such as segmentation~\cite{ros2016synthia,richter2016playing,richter2017playing}, depth/flow estimation~\cite{dosovitskiy2015flownet,mayer2016large,gaidon2016virtual}, object detection~\cite{johnson2016driving}, visual navigation~\cite{savva2019habitat}, and grasping~\cite{coumans2016pybullet}. When there is label scarcity, a popular approach is to resort to training with synthetic images, where full supervision can be obtained at a low cost. This finds applications in label-scarce domains such as robotics and autonomous driving where simulation can play an important role.

However, there are many challenges to train with synthetic images. Models trained on synthetic images often face problems from degraded generalization on the real domain. Such a domain gap is usually caused by limitations on rendering quality, including unrealistic texture, appearance, illumination and scene layout, etc. As a result, networks are prone to overfitting to the synthetic domain with learned representations that differ from those obtained on real images. To this end, domain generalization methods \cite{li2017deeper,pan2018two,yue2019domain} have been proposed to overcome the above domain gaps and improve model generalization on real target domains.

\begin{figure}[t]
\includegraphics[scale=0.35]{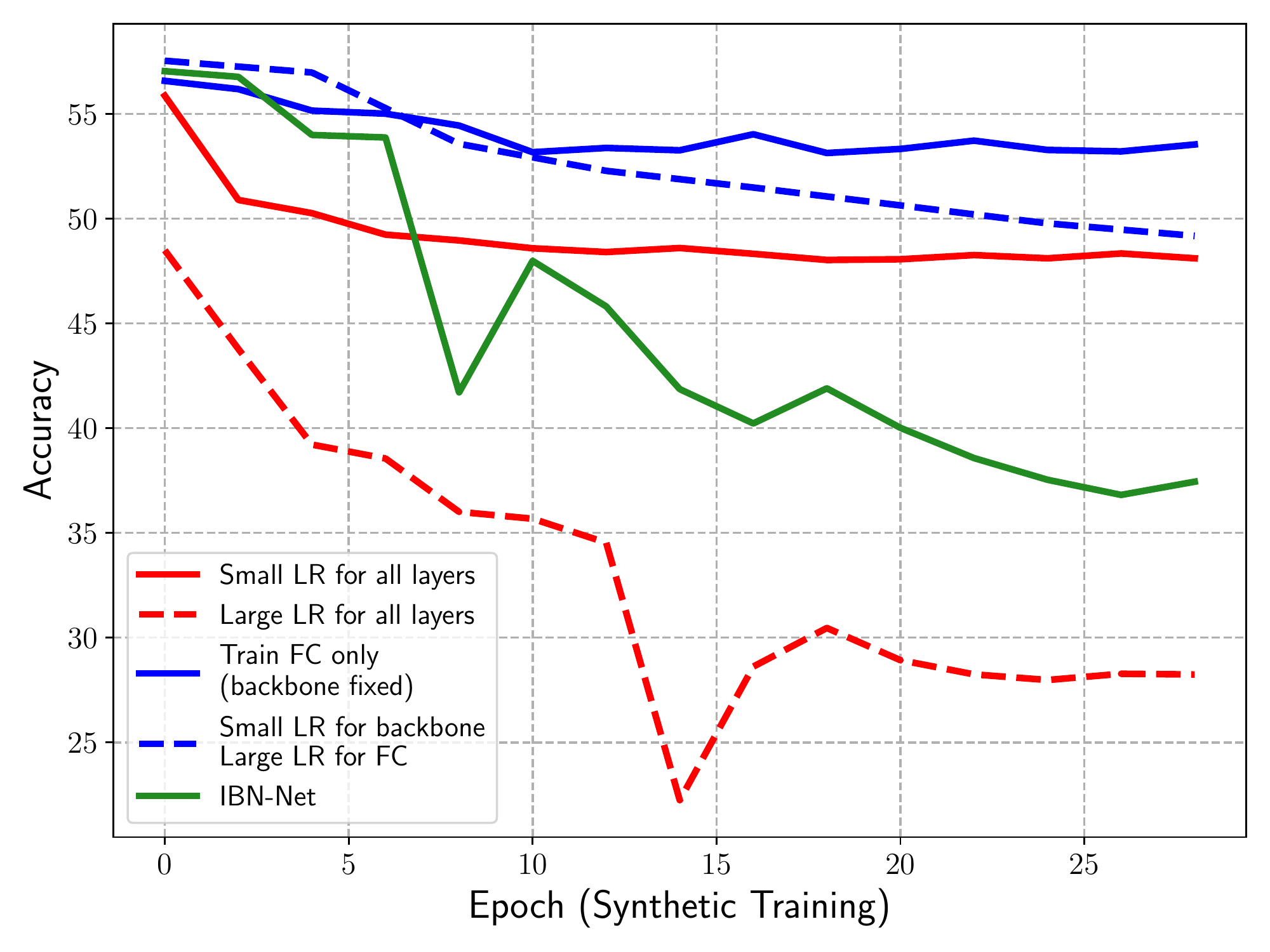}\vspace{-1em}
\centering
\caption{Both heuristic solutions (early stopping, small learning rates, etc.) and recent works (e.g. IBN-Net \cite{pan2018two}) fall in poor generalization in synthetic-to-real transfer learning, which suffers from the huge appearance gap between the source and the target domain. Here, we studied different learning rates (``LR'') or optimization strategies for the backbone and the last fully-connected classification layer (``FC''). All settings start with an ImageNet pre-trained backbone and a randomly initialized classification layer. Please see section \ref{sec:implementations} for experiment details.}
\label{fig:no_lwf}
\vspace{-1.5em}
\end{figure}

Synthetic-to-real transfer learning involves training a model only on synthetic images (source domain) without seeing any real ones, and targets on the generalization performance on unseen real images (target domain).
Recent synthetic-to-real generalization algorithms often start with an ImageNet pre-trained model. To achieve the best generalization performance, it is a common practice to fine-tune the pre-trained model on synthetic images for only a few epochs (i.e. \textbf{early-stopping}) with a \textbf{small learning rate}. Figure~\ref{fig:no_lwf} illustrates the evaluation dynamics of several popular heuristic solutions on the VisDA-17 dataset~\cite{peng2017visda}.
One could clearly see the high performance in early epochs, and the improvements of fine-tuning with a small learning rate (or even a fixed backbone) over training with a large one (red dashed line). Similar behavior exists in recent works (e.g. IBN-Net \cite{pan2018two}).
This observation implies an important clue: all these heuristics try to retain the ImageNet domain knowledge during the synthetic-to-real transfer learning.
It explains why the heuristic solutions in Figure~\ref{fig:no_lwf} work: they allow the classifier to quickly adjust from ImageNet to the task defined by the synthetic images, while preventing the ImageNet pre-trained representations of natural images to be ``washed out'' due to catastrophic forgetting.

Unfortunately, existing solutions (e.g. IBN-Net) still face degraded generalization and are highly dependent on manual selections of training epochs and schedules (learning rates).
Motivated by this open issue, we propose an Automated Synthetic-to-real Generalization (\textbf{ASG}) framework to improve synthetic-to-real transfer learning. This method is automated from two aspects: (1) It stably improves the generalization during transfer learning, avoiding the difficulty of choosing epochs to stop. (2) It automates the complicated tuning of layer-wise learning rates towards better generalization.
The core of our work is the intuition that a good synthetically-trained model should share similar representations with ImageNet-models, and we leverage this intuition as a proxy guidance to search layer-wise training schedules through learning-to-optimize (L2O). 

\subsection*{Summary of Contributions:}
\begin{figure}[h!]
\vspace{-1em}
\includegraphics[width=0.95\linewidth]{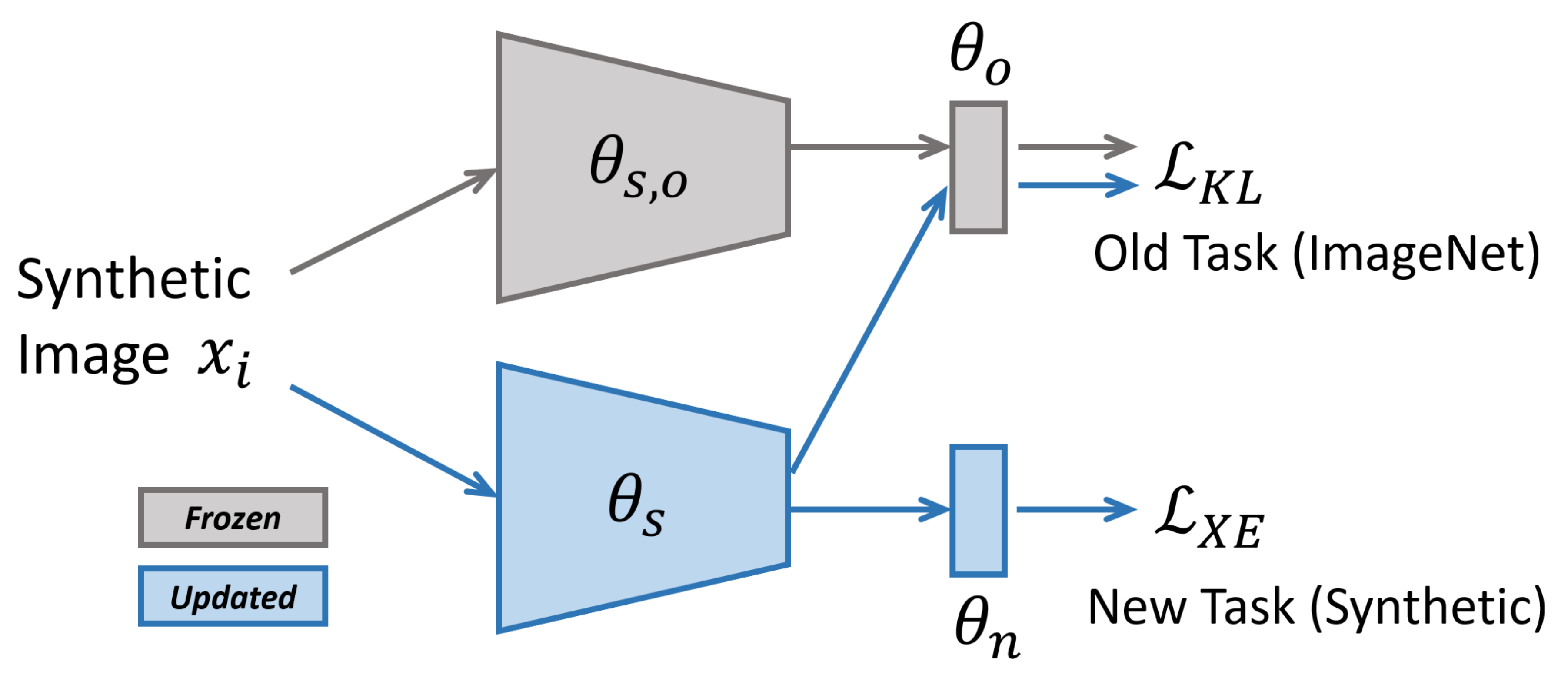}
\centering
\caption{We formulate the synthetic-to-real transfer learning as a lifelong learning problem: training on synthetic images (new task) while still memorizing ImageNet classification (old task), acting as our proxy guidance during the transfer learning.}
\label{fig:lwf}
\end{figure}

\begin{itemize}
    \item We examine the behaviors of various training heuristics, in order to study the role of the ImageNet domain knowledge in synthetic-to-real generalization, which is not thoroughly discussed by the literature to the best of our knowledge.
    \item We provide a novel perspective to address synthetic-to-real generalization, by formulating it as a lifelong learning problem. We enforce the representation similarity between synthetically trained models and ImageNet pre-trained model, and treat their similarity as a proxy guidance of generalization performance. An overall design is illustrated in Figure \ref{fig:lwf}.
    \item We demonstrate that proxy guidance not only dramatically improves the generalization performance, but can also be easily integrated by existing transfer learning frameworks as a simple drop-in module, without requiring any additional training beyond synthetic images. Experiments also prove the cross-task generalizability of our proxy guidance, which magnifies the strength of synthetic-to-real transfer learning.
    \item We design a reinforcement learning based learning-to-optimize (RL-L2O) approach to make the synthetic-to-real generalization practically more convenient, by automating the complicated heuristic designs with layer-wise learning rates. We demonstrate that our RL-L2O method out-performs hand-crafted decisions and learns explainable learning rate strategy.
\end{itemize}

\section{Automated Syn-to-Real Generalization}

In our work, we propose an automated framework to address the synthetic-to-real transfer learning, dubbed Automated Synthetic-to-real Generalization (ASG). We assume an ImageNet pre-trained model as our starting point. Our target is to maximize the performance of the model on a target domain which consists of unseen real images, by utilizing only synthetic images from the source domain.

\subsection{Syn-to-Real Generalization with Proxy Guidance}\label{sec:lwf}
The accessibility to model pre-trained on ImageNet \cite{imagenet_cvpr09} implicitly provides the domain knowledge of real images. As we are transferring a model trained on synthetic data to unseen real images, retaining the ImageNet domain knowledge is potentially beneficial to the generalization. Motivated by this, we force the model to memorize how to capture the representation learned from ImageNet while training on synthetic images, to maintain both the domain knowledge on real images and task-specific information provided by the synthetic data.

We start with an ImageNet pre-trained model $\mathcal{M}$,
and formulate our transfer learning as a life-long learning problem: training on synthetic images as the new task while still memorizing the old ImageNet classification task.
While updating the model $\mathcal{M}$ with synthetic images, we also keep a copy of the original ImageNet pre-trained model $\mathcal{M}_o$ which is frozen during the training. In addition to the cross-entropy loss $\mathcal{L}_{\mathrm{XE}}$ calculated on the synthetic dataset, we also forward the synthetic images through $\mathcal{M}_o$ and minimize the KL divergence $\mathcal{L}_{\mathrm{KL}}$ between the output of $\mathcal{M}_o$ and $\mathcal{M}$. Formally, we leverage the minimization of $\mathcal{L}_{\mathrm{KL}}$ as a proxy guidance during our transfer learning process:

{\small
\vspace{-0.3cm}
\begin{align}
    \theta_s^*, \theta_n^* &\leftarrow \underset{\theta_s, \theta_n}{\mathrm{\argmin}} (\mathcal{L}) \\
    \mathcal{L} &= \mathcal{L}_{\mathrm{XE}} + \lambda\mathcal{L}_{\mathrm{KL}} \label{eq:lwf} \\
    \mathcal{L}_{\mathrm{XE}} &= - \frac{1}{N_B} \sum_{i=1}^{N_B} \mathbf{y}_{i} \mathrm{log}(\mathcal{M}(\mathbf{x}_{i}, \theta_s, \theta_n)) \label{eq:xe} \\
    \mathcal{L}_{\mathrm{KL}} &= - \frac{1}{N_B} \sum_{i=1}^{N_B} \mathcal{M}_o(\mathbf{x}_{i}, \theta_{s,o}, \theta_o) \mathrm{log}(\mathcal{M}(\mathbf{x}_{i}, \theta_s, \theta_o)) \label{eq:kl}
\end{align}
\vspace{-1em}
}

Here, $\lambda$ is a balancing factor that controls how much ImageNet domain knowledge the model should retain. $\theta_n$ denotes the parameters for the synthetic-to-real transfer learning $\mathcal{L}_{\mathrm{XE}}$ (i.e. the classifier layers for the new task), $\theta_o$ denotes the parameters for ImageNet classifier which will output the predicted probabilities on the ImageNet domain. $\theta_s$ denotes the parameters for the feature extractor (a.k.a. backbone) updated for the new tasks, and $\theta_{s,o}$ denotes the parameters for the feature extractor which is frozen with ImageNet pre-trained weights. $\theta_{s}$ and $\theta_{s,o}$ share the same structure. $N_B$ is the current batch size, $\mathbf{x}_{i}$, and $\mathbf{y}_{i}$ are sample and ground truth from the new task in the current batch. This synthetic-to-real transfer learning with proxy guidance is illustrated in Figure \ref{fig:lwf}. The new task and the old ImageNet task are jointly optimized during the training.

\textbf{Cross-task proxy guidance}: It is important to note that, the new task is not necessarily limited to be also for the image classification purpose. For some models in semantic segmentation (e.g. ResNet based FCN \cite{long2015fully}), a pixel-wise $\mathcal{L}_{\mathrm{XE}}$ provides a much denser supervision than the image-wise $\mathcal{L}_{\mathrm{KL}}$ in Eq. \ref{eq:kl}. To spatially balance $\mathcal{L}_{\mathrm{XE}}$ and $\mathcal{L}_{\mathrm{KL}}$, we also make $\mathcal{L}_{\mathrm{KL}}$ denser by applying it on cropped feature map patches:
\begin{equation}
    \footnotesize
    \mathcal{L}_{\mathrm{KL}}^{\mathrm{dense}} = - \frac{1}{N_B} \frac{1}{N} \sum_i^{N_B} \sum_j^{N} \mathcal{M}_o(\mathbf{x}_{i,j}, \theta_{s,o}, \theta_o) \mathrm{log}(\mathcal{M}(\mathbf{x}_{i,j}, \theta_s, \theta_o)). \label{eq:kl_dense}
\end{equation}
Here, $\mathbf{x}_{i,j}$ $(j=1,\cdots,N)$ are cropped patches from $\mathbf{x}_{i}$. Later in section \ref{sec:exp_seg_transfer} we will demonstrate that this formulation also works well for cross-task training.

\subsection{Automate LR Selection via Learning-to-Optimize}

As observed in Figure \ref{fig:no_lwf}, different convolution blocks contribute differently to the generalizability. This leads to a question: \textit{does different layers in a deep network require different training strategy towards optimal synthetic-to-real generalization performance during the transfer learning?}

To avoid manually tuning the hyperparameters, we propose a reinforcement learning based learning-to-optimize (RL-L2O) framework to automatically adjust the learning rates for layers.
In the RL-L2O framework, we aim to learn a parameterized policy $\pi$ to dynamically control the learning rates given the training statistics of our model $\mathcal{M}$ during transfer learning.

Generally, the goal of the reinforcement learning algorithm is to learn a policy $\pi^*$ that maximizes the total expected reward $r$ over time. More precisely,
\begin{equation}
    \pi^* = \argmax_\pi \mathds{E}_{\bm{s}_0, \bm{a}_0, \bm{s}_1, ..., \bm{s}_T} \begin{bmatrix} \sum\limits_{t=0}^T r_t \end{bmatrix}
\end{equation}
where the expectation is taken over the sequence of states (or observations) and actions. In short, an action $\bm{a}_t$ produced by $\pi$ will update the learning rates for $\mathcal{M}$ in the RL-L2O framework. A state $\bm{s}_t$ contains optimization related statistics of the model $\mathcal{M}$ during the transfer learning, and the reward $r_t$ measures how well the optimization performs.

\textbf{Design of Optimization Coordinates:} One challenge in applying reinforcement learning in our setting is that we want to be able to control the training schedules of a deep network of up to a hundred layers (ResNet-101), each of them requiring an action from our policy.
As layers may have strong correlations during the optimization \cite{ghiasi2018dropblock}, the policy may fall into sub-optimal solutions in this large scale action space. To avoid this difficulty and simplify our policy training, we leverage the underlying structures in current deep networks. Specifically, layers in $\mathcal{M}$ with similar input resolution will be grouped into a block, named as an \textit{optimization coordinate}.
Taking the ResNet family as an example, we group layers into a new coordinate whenever the feature map resolution is reduced. This grouping strategy keeps the action space of the policy small, and speeds-up the L2O training.

\textbf{Design of Action Space:} Intuitively, our policy could directly output learning rate for each coordinate. However, the model $\mathcal{M}$ could be very sensitive to the learning rate (as observed in Figure \ref{fig:no_lwf}), and the learning rate usually resides in a small value range (e.g. $10^{-4} \sim 10^{-3}$). Directly predicting the value of the learning rate could be very unstable. Instead, we propose a learning rate scaling factor as the action. We first provide the policy a base learning rate $\eta_{\mathrm{base}}$. In the following steps, $\pi$ outputs discrete coordinate-wise learning rate scale factors as its actions $\bm{a}_t = [a_{1,t}, ..., a_{C,t}]$ where $C$ is the number of optimization coordinates in $\mathcal{M}$. We formulate $\bm{a}_t$ as categorical actions, where each learning rate scale factor $a_{c,t} \in [0, 0.1, 0.2, ..., 0.9, 1]$. The learning rate for each coordinate is set to be $\eta_{c,t} = a_{c,t} \cdot \eta_{\mathrm{base}}$, and we leverage the gradients and momentums calculated by stochastic gradient descent (SGD) \cite{rumelhart1986learning} to update the parameters in $\mathcal{M}$.

\begin{figure}[h!]
\centering
\includegraphics[scale=0.38]{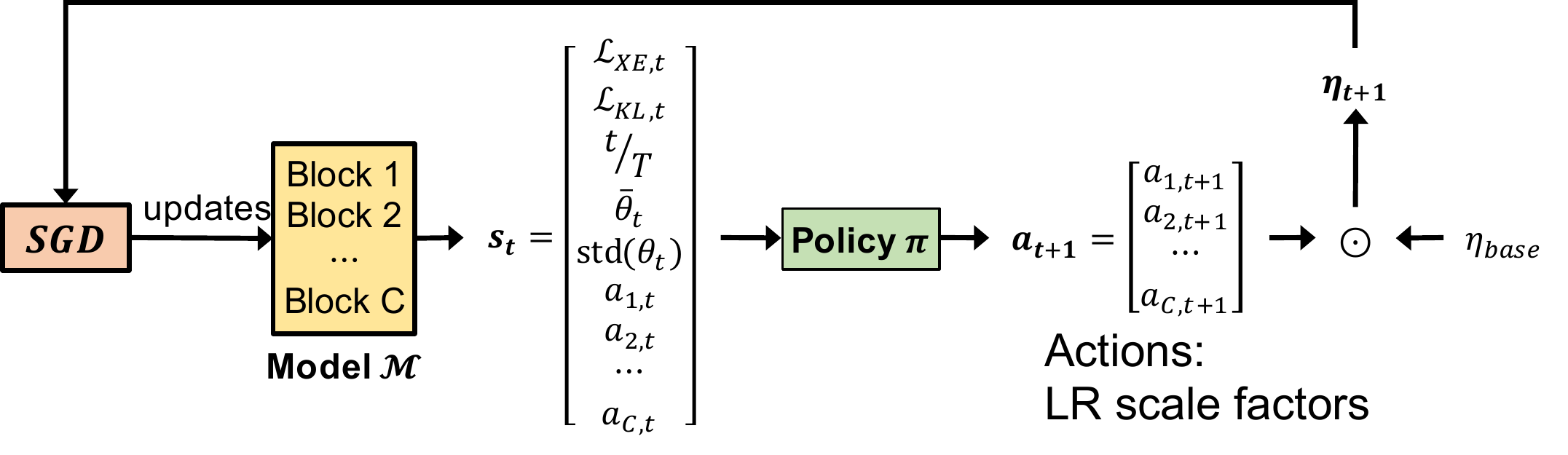}\vspace{-2.5em}
\caption{Workflow of the proposed L2O framework. $\bm{a}_t = [a_{1,t}, ..., a_{C,t}]^T$ is the learning rate scale factor for the coordinates, and $\eta$ indicates the learning rate. $\odot$ is dot product.}
\label{fig:l2o_lstm}\vspace{-0.5em}
\end{figure}

\begin{figure}[h!]
\includegraphics[scale=0.35]{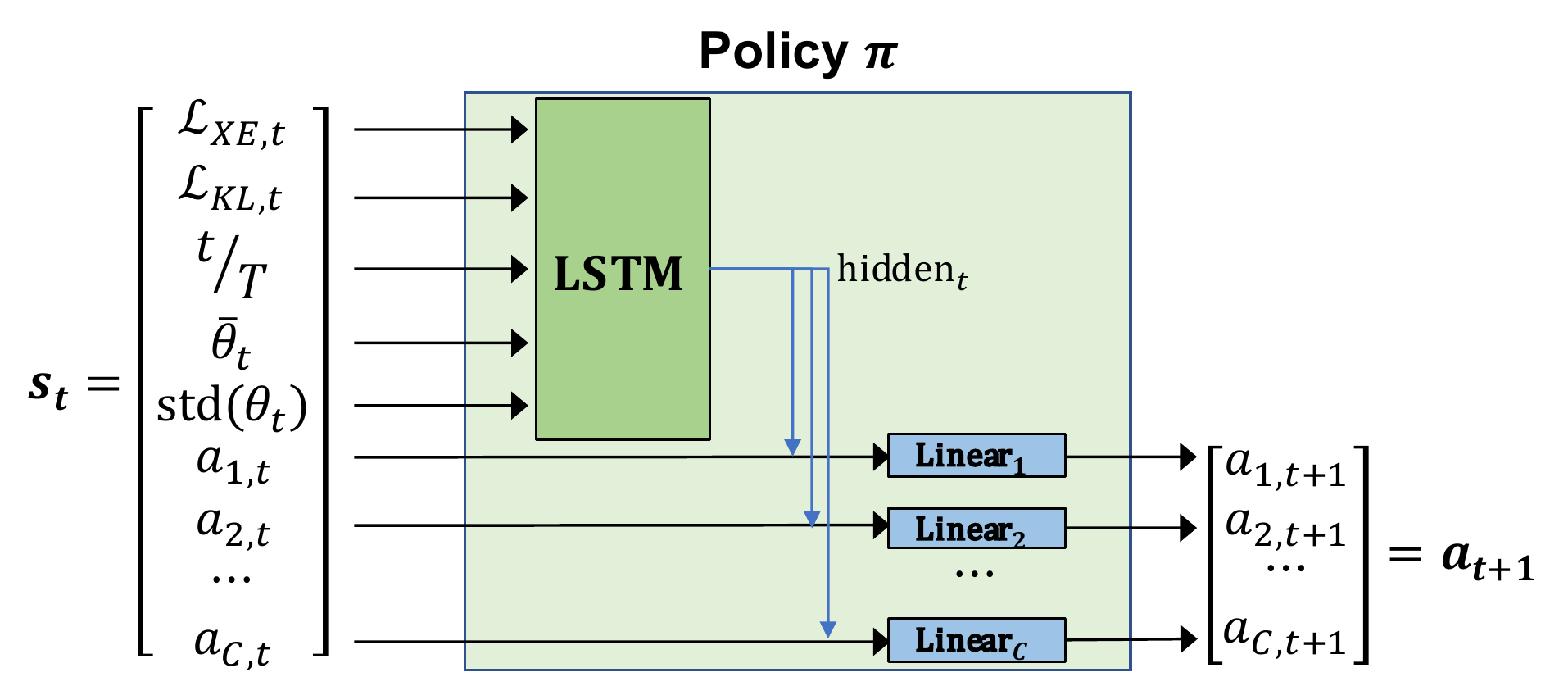}\vspace{-1em}
\centering
\caption{Architecture of the policy network.}
\label{fig:policy}\vspace{-1em}
\end{figure}

\textbf{Design of Observation Space and Reward:} At each step, the state (observation) $\bm{s}_t$ for $\pi$ includes: current $\mathcal{L}_{\mathrm{XE},t}$ (Eq. \ref{eq:xe}) and $\mathcal{L}_{\mathrm{KL},t}$ (Eq. \ref{eq:kl}, Eq. \ref{eq:kl_dense}), the training progress of $\mathcal{M}$ (i.e. $\ddfrac{t}{T}$, where $T$ equals to total training steps (i.e., ``total epochs''$\times$``iterations per epoch'')), the mean and standard deviation of the weights of the classifier ($\bar{\theta}_{n,t}$ and $\mathrm{std}(\theta_{n,t})$), and finally the scale factors from the last step $\bm{a}_{t-1}$. The policy learning is guided by reward $r_t = \mathcal{L}_{t-1} - \mathcal{L}_t$.

\textbf{Policy Training:} We update our $\mathrm{LSTM}$ policy $\pi$ via the REINFORCE algorithm \cite{williams1992simple} to minimize:
\begin{equation}
    \mathcal{L}_\pi = -\frac{1}{U} \sum\limits_{t \in \mathrm{U}} r_t \cdot \mathrm{log}(p_\pi(\bm{a}_t | \bm{s}_t)),
\end{equation}
where $U$ is the unroll length for $\mathrm{LSTM}$. Algorithm \ref{algo:L2O} illustrate the procedure of our RL-L2O framework.

\vspace{-1em}
\begin{algorithm2e}[h!]
	\textbf{Input:} base learning rate $\eta_{\mathrm{base}}$, parameters $\theta_{n,0}, \theta_{s,0}$, hidden state $\bm{h}_0 = \bm{0}$, policy $\pi$, unroll length $U$, total training steps $T$. \\
	Calculate $\mathcal{L}_0, \mathcal{L}_{\mathrm{XE}, 0}, \mathcal{L}_{\mathrm{KL}, 0}$ for $\theta_{n,0}, \theta_{s,0}$\\
	Initialize $\mathrm{storage}$\\
		\For {$t=0,\dots,T - 1$}{
                $\mathrm{prob(}\bm{a}_{t+1})$, $\bm{a}_{t+1}$, $h_{t+1}$ = $\pi$($\mathcal{L}_{\mathrm{XE}, t}, \mathcal{L}_{\mathrm{KL}, t}, \ddfrac{t}{T}, \bar{\theta}_{n,t}, \mathrm{std}(\theta_{n,t}), \bm{a}_{t}, h_{t}$)\\
            ($\theta_{n,t+1}, \theta_{s,t+1}$) = SGD($\nabla \mathcal{L}_{t}, \bm{a_{t+1}}, \eta_{\mathrm{base}}, \theta_{n,t}, \theta_{s,t}$) \\
                Calculate $\mathcal{L}_{t+1}, \mathcal{L}_{\mathrm{XE}, {t+1}}, \mathcal{L}_{\mathrm{KL}, {t+1}}$ for $\theta_{n,{t+1}}, \theta_{s,{t+1}}$\\
            $r_{t+1} = \mathcal{L}_{t} - \mathcal{L}_{t+1}$\\
            $\mathrm{storage.append(prob(}\bm{a}_{t+1}$), $r_{t+1})$\\
		\If{$(t+1) \% U == 0$}{
			$\pi = \mathrm{REINFORCE(}\pi, \mathrm{storage)}$\\
			Initialize $\mathrm{storage}$
		}
		\Return{final learned policy $\pi$.}
	}
	\caption{RL-L2O: policy ($\pi$) learning to control group-wise learning rates.}
	\label{algo:L2O}
\end{algorithm2e}
\vspace{-1em}

Once we obtained the learned policy $\pi$, we then freeze and apply it to the synthetic-to-real transfer learning of $\mathcal{M}$ together with SGD, as illustrated in Figure \ref{fig:l2o_lstm}.

\section{Experiments}
\subsection{Datasets}\label{sec:dataset}
\textbf{VisDA-17} \cite{peng2017visda} We perform ablation study on the VisDA-17 image classification benchmark. The VisDA-17 dataset provides three subsets (domains), each with the same 12 object categories. Among them, the training set (source domain) is collected from synthetic renderings of 3D models under different angles and lighting conditions, whereas the validation set (target domain) contains real images cropped from the Microsoft COCO dataset \cite{lin2014microsoft}.

\textbf{GTA5} \cite{richter2016playing} is a vehicle-egocentric image dataset collected in a computer game with pixel-wise semantic labels. It contains 24,966 images with a resolution of 1052$\times$1914. There are 19 classes that are compatible with the Cityscapes dataset.

\textbf{Cityscapes} \cite{Cordts2016Cityscapes} contains urban street images taken on a vehicle from some European cities. There are 5,000 images with pixel-wise annotations. The images have a resolution of 1024$\times$2048 and are labeled into 19 semantic categories.

\subsection{Implementation}\label{sec:implementations}

\textbf{Image classification}: For VisDA-17, we choose ResNet-101 \cite{he2016deep} as the backbone, and one fully-connected layer as the classifier. Backbone is pre-trained on ImageNet \cite{imagenet_cvpr09}, and then fine-tuned on source domain, with learning rate = $1\times 10^{-4}$, weight decay = $5\times 10^{-4}$, momentum = 0.9, and batch size = 32. The model is trained for 30 epochs and $\lambda$ for $\mathcal{L}_{\mathrm{KL}}$ is set as $0.1$. In section~\ref{sec:asg_clf}, we will additionally study how to choose $\lambda$.

\textbf{Semantic segmentation}: We study both FCN with ResNet-50 and FCN with VGG-16 \cite{long2015fully}.
Backbones are pre-trained on ImageNet. Our learning rate is $1\times 10^{-3}$, weight decay is $5\times 10^{-4}$, momentum is 0.9, and batch size is six. We crop the images into patches of 512$\times$512 and train the model with multi-scale augmentation (0.75 $\sim$ 1.25) and horizontal flipping. The model is trained for 50 epochs, and $\lambda$ for $\mathcal{L}_{\mathrm{KL}}$ is set as $75$. Note that $\lambda$ in segmentation is considerably larger since $\mathcal{L}_{\mathrm{XE}}$ is a pixel-wise dense loss.

\textbf{RL-L2O policy}: We set the learning rate for policy training as $0.5$.
The size of the hidden state vector $\bm{h}$ is set to 20, and the unroll length $U = 5$. We train $\pi$ for 50 epochs. For the ResNet family, we follow the convention \cite{he2016deep} to group the layers into $C = 7$ coordinates: $\mathrm{conv1}, \mathrm{bn1},\mathrm{conv2}, \mathrm{conv3}, \mathrm{conv4}, \mathrm{conv5}$, and the $\mathrm{classifier}$.
For VGG-16 \cite{long2015fully}, we also group the layers into $C = 7$ coordinates: $\mathrm{conv1}, \mathrm{conv2}, \mathrm{conv3}, \mathrm{conv4}, \mathrm{conv5}, \mathrm{conv6\&7}$, and the remaining $\mathrm{projection}\_\mathrm{upsampling}$ layers.

\textbf{Proxy guidance}: For all backbones we studied (ResNet-50, ResNet-101, and VGG-16), we forward the feature maps extracted by group $\mathrm{conv5}$ into the ImageNet classifier (parameterized by $\theta_o$) to calculate $\mathcal{L}_{\mathrm{KL}}$.

\subsection{ASG for Image Classification}\label{sec:asg_clf}
We first perform the ablation studies on the VisDA-17 image classification task\footnote{There is no previous synthetic-to-real transfer work on VisDA-17 classification task, only domain adaptation works.}.

\begin{figure}[h!]
\includegraphics[scale=0.35]{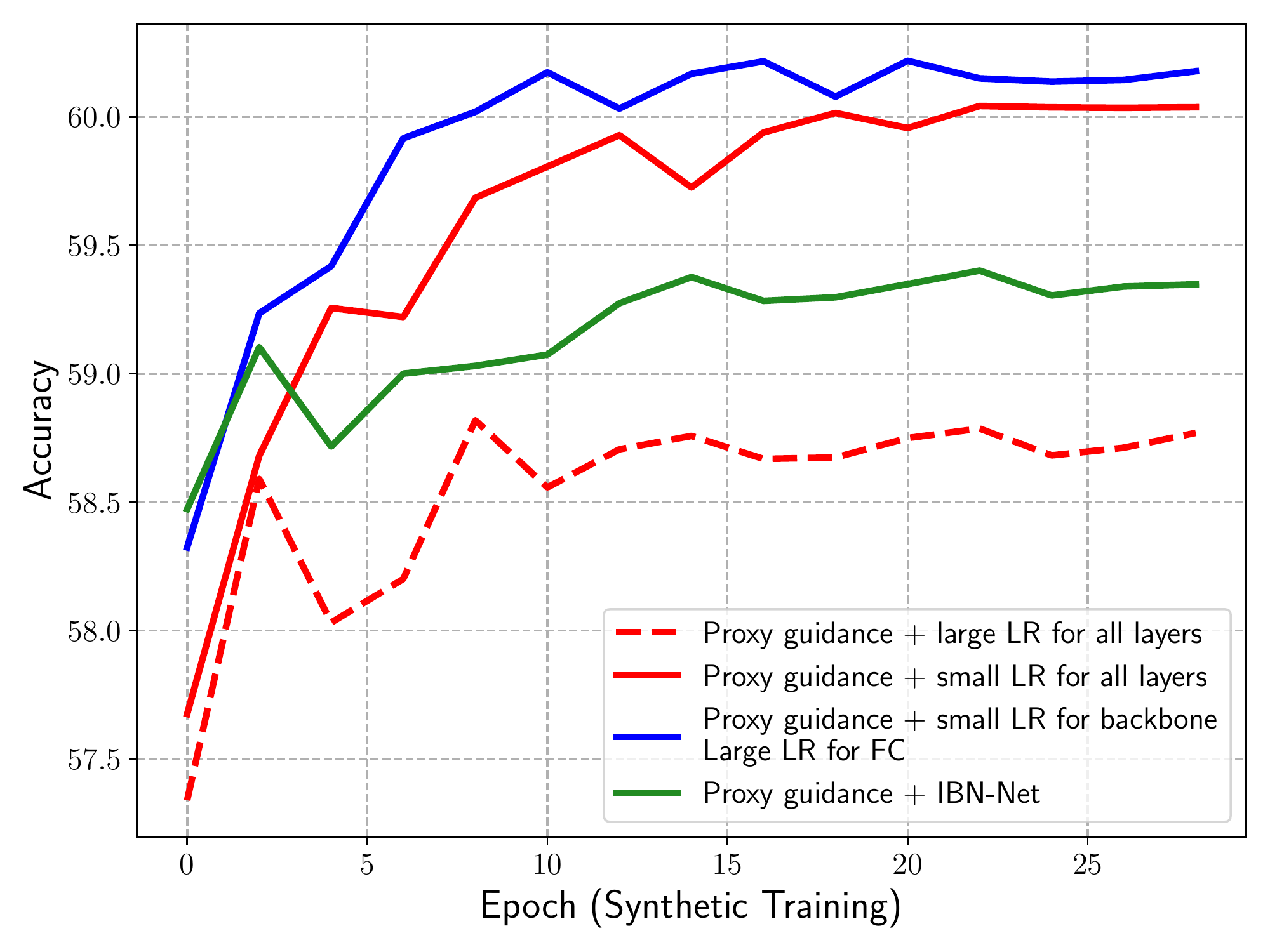}\vspace{-1em}
\centering
\caption{The degraded generalization during the synthetic-to-real transfer learning (studied in Figure \ref{fig:no_lwf}) can be solved by forcing the model to retain the ImageNet domain knowledge via our proxy guidance\footnote[2]. Task: ResNet-101 VisDA-17 Classification. $\lambda = 0.1$.}
\label{fig:lwf_curve}
\end{figure}

\textbf{Generalization with Proxy Guidance.} To evaluate the effect of our proxy guidance, we apply our $\mathcal{L}_{\mathrm{KL}}$ loss on different learning rate settings we studied in Figure \ref{fig:no_lwf}. As demonstrated in Figure \ref{fig:lwf_curve}, once we force the model to memorize the ImageNet domain knowledge, we achieve stably increasing and eventually better generalization performance for each setting we explored in Figure \ref{fig:no_lwf}.
The relative ranking still holds among the different learning rate settings, while the degraded generalizability is addressed. Early stopping is no longer needed, as models enjoy improved generalization given sufficient training epochs.
This ablation study validates the contribution of retaining the ImageNet domain knowledge during the synthetic-to-real transfer learning. It is also worth noting that our proxy guidance can be also applied to different networks (e.g. the IBN-Net \cite{pan2018two}, green line in Figure \ref{fig:lwf_curve}), which demonstrate the easy integration of our approach as a simple drop-in module with existing synthetic-to-real generalization works, without requiring any additional training beyond synthetic images.

\footnotetext[2]{We could not utilize the proxy guidance when the backbone is fixed (``Train FC Only'' blue dashed curve in Figure \ref{fig:no_lwf}). The $\mathcal{L}_{\mathrm{KL}}$ is always zero in this case as the group $\mathrm{conv5}$ is not updated.}

Moreover, a vital conclusion from Figure \ref{fig:lwf_curve} is that, only reporting the (final) performance as a number is far from sufficient for analyzing and comparing synthetic-to-real transfer learning methods. Instead, the curve of the target performance during training can better demonstrate how well a model's generalizability is. Meanwhile, a stably increasing training curve implies that, the model is both better leveraging synthetic images and retaining ImageNet domain knowledge, instead of overfitting on synthetic appearance and leaving the domain gap an open issue.

\textbf{How to choose $\lambda$:} We also study the effect of different strengths of the proxy guidance loss $\mathcal{L}_{\mathrm{KL}}$ by adjusting $\lambda$ in Equation \ref{eq:lwf} for a ResNet-101 model trained with a small learning rate for the backbone and a large one for the classification layer (blue line in Figure \ref{fig:lwf_curve}). In Table \ref{table:lambda}, we adjust $\lambda$ in a wide range from 0.01 to 1. While we obtain the best generalization accuracy with $\lambda = 0.1$, we can see that our proxy guidance is very robust to different strength of $\mathcal{L}_{\mathrm{KL}}$. Therefore, choosing $\lambda$ is much easier than tuning hyperparameters in heuristic solutions like epochs.

\begin{table}[h!]\vspace{-1em}
\caption{Ablation of $\lambda$ for the proxy guidance loss $\mathcal{L}_{\mathrm{KL}}$. Model: ResNet-101. Task: VisDA-17 Classification.}\vspace{0.3em}
\centering
\begin{tabular}{cccccc}
\toprule
$\lambda$ & 0.01 & 0.05 & 0.1 & 0.5 & 1 \\ \midrule
Accuracy (\%) & 58.9 & 59.4 & \textbf{60.1} & 58.5 & 59.7 \\
\bottomrule
\end{tabular}
\label{table:lambda}
\end{table}

\begin{figure}[h!]
\includegraphics[scale=0.35]{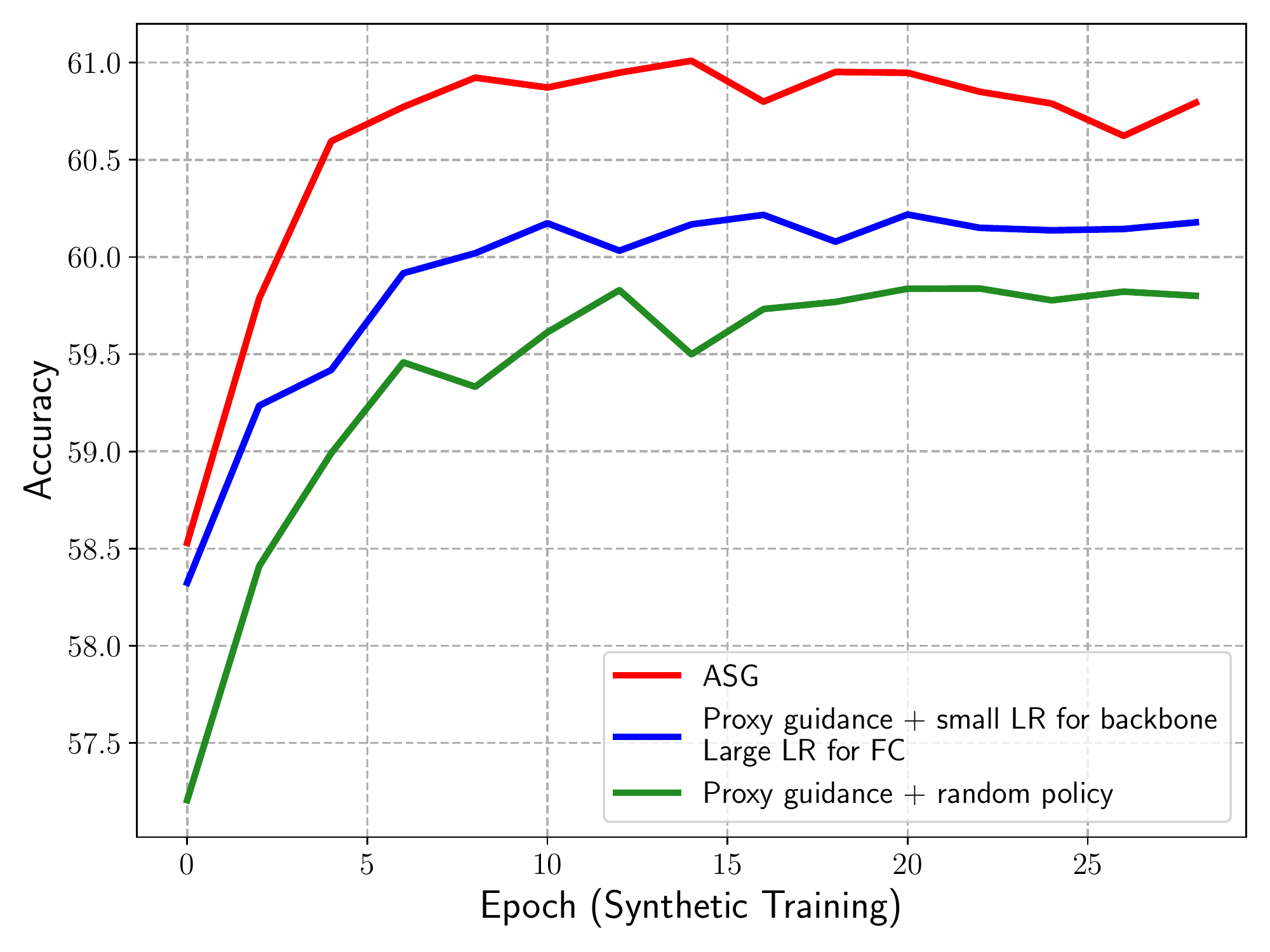}\vspace{-1em}
\centering
\caption{Our RL-L2O framework can out-perform both the random policy and a carefully hand-tuned learning rate strategy. All three settings include $\mathcal{L}_{\mathrm{KL}}$ with the same $\lambda = 0.1$ during training. Model: ResNet-101. Task: VisDA-17 Classification.}
\label{fig:l2o_cls_acc}
\end{figure}

\textbf{Automated Syn-to-Real Generalization.} We next evaluate the performance of our RL-L2O framework. Specifically, we want to make sure the policy learned by our RL-L2O can perform better than both the random policy and the best hand-tuned learning rate policy we explored in Figure \ref{fig:lwf_curve}. A random policy means that the controller will always randomly pick an action as the learning rate scale factor. In all these three settings we start from the same base learning rate $\eta_{\mathrm{base}} = 1\times 10 ^ {-4}$. Figure \ref{fig:l2o_cls_acc} demonstrates that, although the hand-tuned learning rate strategy is better than a random policy, RL-L2O can still out-perform it (blue line).

\textbf{Additional Ablation Study on VisDA-17.} We conduct additional ablation studies on VisDA-17 to further analyze the learning behaviors of ASG. Specifically, as both the proxy guidance and the RL-L2O frameworks are motivated to carefully preserve the ImageNet representations while targeting updates from the new tasks on synthetic data, it is interesting and important to connect the relation between the level of retained ImageNet knowledge and the synthetic-to-real generalization. In our experiment, we compute ImageNet validation accuracy as well as the generalization performance on Visda-17 target domain for the classification task.

Table \ref{table:retain_imagenet} demonstrates two conclusions: 1) Heuristic solutions that retain more ImageNet domain knowledge achieve higher synthetic-to-real generalization (\#3 versus \#1), i.e., using hand-crafted small learning rates to prevent the ImageNet pre-trained representations of natural images from being ``washed out'' due to catastrophic forgetting; 2) By leveraging Proxy Guidance, the generalization performance on VisDA-17 is dramatically improved, while the ImageNet accuracy is also maintained with almost no drop. It is interesting that Proxy Guidance leads to learned model parameters that achieve high accuracy simultaneously on both ImageNet and VisDA-17. In contrast, naively freezing the backbone and only fine-tuning the classifier layer (``Oracle'' \#5) results in inferior synthetic-to-real generalization despite high ImageNet performance.

\begin{table}[h!]
\caption{Our Proxy Guidance improves the synthetic-to-real generalization (Visda-17) by retaining the ImageNet domain knowledge. Learning rate (LR) settings were studied in Figure \ref{fig:no_lwf} and \ref{fig:lwf_curve}. FC:  the last fully-connected classification layer. Top1 accuracies are in percentage (\%). Model: ResNet-101.}\vspace{0.3em}
\centering
{\small
\begin{tabular}{cccc}
\toprule
\# & Model & VisDA-17 & ImageNet \\ \midrule
1. & Large LR for all layers & 28.2 & 0.8 \\
2. & + our Proxy Guidance & 58.7 (+30.5) & 76.2 (+75.4) \\ \midrule
\multirow{2}{*}{3.} & Small LR for backbone & \multirow{2}{*}{49.3} & \multirow{2}{*}{33.1} \\
& and large LR for FC & & \\
4. & + our Proxy Guidance & 60.2 (+10.9) & 76.5 (+43.4) \\ \midrule
5. & Oracle on ImageNet\footnote[3] & 53.3 (+4.0) & \textbf{77.4} \\
6. & ROAD \cite{chen2018road} & 57.1 (+7.8) & \textbf{77.4} \\
7. & Vanilla L2 distance & 56.4 (+7.1) & 49.1 \\
8. & SI \cite{zenke2017continual} & 57.6 (+8.3) & 53.9 \\ \midrule
9. & ASG (ours) & \textbf{61.1} & 76.7 \\ \bottomrule
\end{tabular}
}
\label{table:retain_imagenet}
\end{table}

\footnotetext[3]{Oracle is obtained by freezing the ResNet-101 backbone while only training the last new fully-connected classification layer on the Visda-17 source domain (the FC layer for ImageNet remains unchanged). We use the PyTorch official model of ImageNet pre-trained ResNet-101.}

In addition, we compare ASG with several other lifelong learning algorithms, including both feature-level $\ell_2$ regularization \cite{chen2018road} and weight-level importance-reweighted $\ell_2$ constraints \cite{zenke2017continual}. Row \#5$\sim$8 in Table \ref{table:retain_imagenet} shows that although the three comparing methods indeed retain ImageNet domain knowledge while improving over the baseline (49.3\%), they are not performing as well as the proxy guidance (60.2\%) under the same LR policy.

\subsection{ASG for Semantic Segmentation}\label{sec:exp_seg_transfer}

We also conduct comprehensive experiments to evaluate the synthetic-to-real generalization performance of ASG on the semantic segmentation task. In particular, we treat GTA5 as the synthetic source domain and train segmentation models on it. We then treat the Cityscapes validation/test sets as target domains where we directly evaluate the segmentation performance of the synthetically trained models.

\begin{figure}[h!]
\includegraphics[scale=0.35]{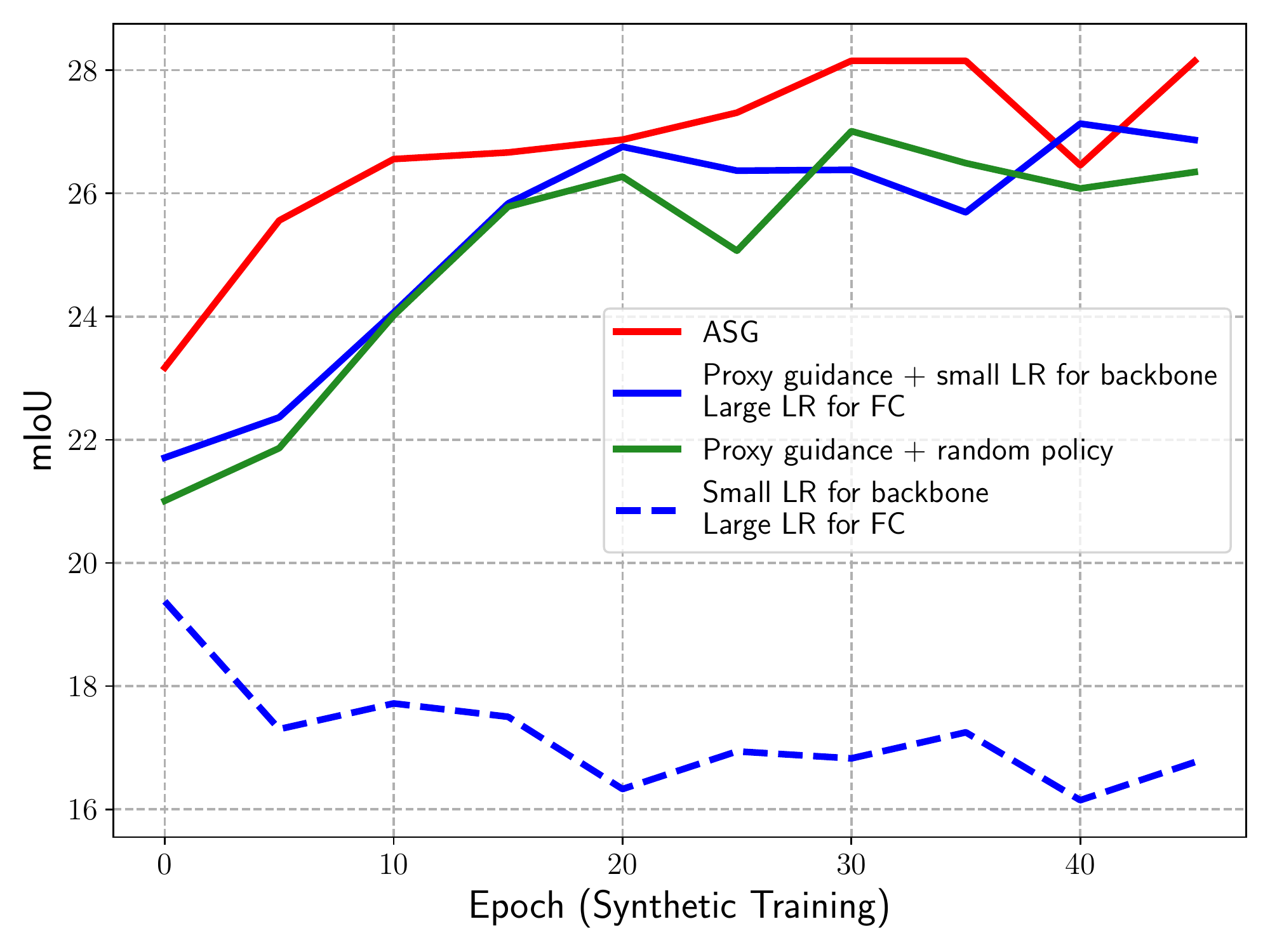}\vspace{-1em}
\centering
\caption{Dynamics of evaluation accuracy with training epochs. Models are trained on GTA5 and directly tested on the Cityscapes validation set. We use FCN-VGG16 as the backbone for segmentation models. In addition, $\mathcal{L}_{\mathrm{KL}}$ in all comparing methods share the same parameter $\lambda = 75$ during synthetic source training.}
\label{fig:seg_vgg16}
\end{figure}

Figure \ref{fig:seg_vgg16} shows the dynamics of evaluation accuracy on the Cityscapes validation set. Again, ASG demonstrates significantly improved generalization performance on semantic segmentation over naive synthetic training. In addition, integrating proxy guidance with RL-L2O also consistently outperforms baselines where proxy guidance is integrated with other policy strategies. Note that in this case, both $\theta_o$ and $\mathcal{L}_{\mathrm{KL}}$ are oriented to the classification task, while $\theta_n$ and $\mathcal{L}_{\mathrm{XE}}$ designed for segmentation. This showcases the ability of ASG to generalize across different tasks.

In Table \ref{table:gta5transfer}, we compare our method with prior domain generalization methods for semantic segmentation. One can see that ASG achieves the best performance gain. Among the comparing methods, IBN-Net \cite{pan2018two} improves domain generalization by fine-tuning the mixed IN-BN residual building blocks, while \cite{yue2019domain} transfers the styles from images in ImageNet to synthetic images. It is worth noting that \cite{yue2019domain} requires ImageNet images during training and implicitly leverages ImageNet label information (i.e. ``Auxiliary Domains'') which brings potential advantages. In contrast, our method requires minimum extra information without using any additional images or labels, therefore can be conveniently applied to existing frameworks as a drop-in training strategy.

\begin{table}[h!]
\vspace{-1em}
\caption{Comparison to prior methods on domain generalization for semantic segmentation (GTA5$\rightarrow$ Cityscapes).}\vspace{0.3em}
\footnotesize
\centering
\begin{tabular}{cccc}
\toprule
Methods & Model & mIoU \% & mIoU $\uparrow$ \% \\ \midrule
No Adapt & \multirow{2}{*}{FCN-Res50} & 22.17 & \multirow{2}{*}{7.47} \\
IBN-Net \yrcite{pan2018two} & & 29.64 &  \\ \midrule
No Adapt & \multirow{2}{*}{FCN-Res50} & 32.45 & \multirow{2}{*}{4.97} \\
Yue et al. \yrcite{yue2019domain} & & 37.42 &  \\ \midrule
No Adapt & \multirow{2}{*}{FCN-Res50} & 23.29  & \multirow{2}{*}{\textbf{8.60}} \\
Ours & & 31.89  & \\ \midrule[0.8pt]
No Adapt & \multirow{2}{*}{FCN-VGG16} & 29.81 & \multirow{2}{*}{6.3} \\
Yue et al. \yrcite{yue2019domain} & & 36.11 &  \\ \midrule
No Adapt & \multirow{2}{*}{FCN-VGG16} & 19.89  & \multirow{2}{*}{\textbf{11.58}} \\
Ours & & 31.47  & \\ \bottomrule
\end{tabular}\label{table:gta5transfer}
\end{table}

\begin{figure}[h!]
\includegraphics[scale=0.35]{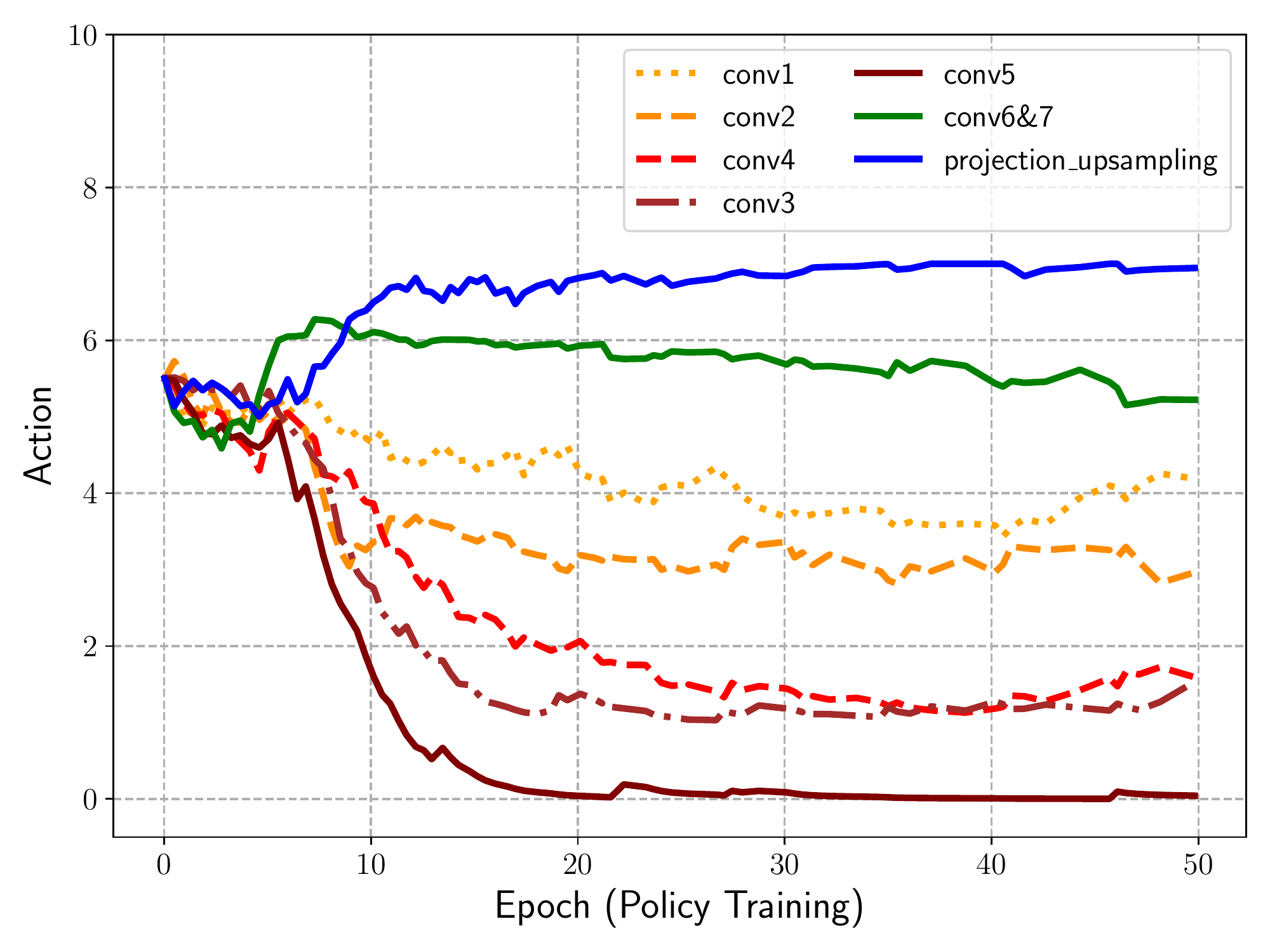}\vspace{-1em}
\centering
\caption{Action behavior of our RL-L2O framework during the policy training for $\mathcal{M}$ = FCN-VGG16 for the GTA5$\rightarrow$Cityscapes segmentation transfer learning. Categorical actions are smoothed for better visualization purpose. Actions of $[0, 1, \cdots, 10]$ indicate learning rate scale factors $[0, 0.1, \cdots, 1.0]$.}
\label{fig:l2o_behavior}
\end{figure}

\textbf{Policy Behaviors.} Figure \ref{fig:l2o_behavior} shows clear and explainable behavior patterns of our policy for FCN-VGG16 on the segmentation task.
In FCN-VGG16, groups $\mathrm{conv1-5}$ belong to the ImageNet pre-trained backbone, while $\mathrm{conv6\&7}$ and the remaining $\mathrm{projection}\_\mathrm{upsampling}$ layers act as the classifier for the dense predictions. The feature map captured by $\mathrm{conv5}$ is forward into $\theta_o$ to calculate $\mathcal{L}_{\mathrm{KL}}$. As $\mathrm{conv5}$ is close to the calculation of $\mathcal{L}_{\mathrm{KL}}$, fixing $\mathrm{conv5}$ (i.e. selecting action = 0 which represents the learning rate scale factor = 0) can effectively minimize $\mathcal{L}_{\mathrm{KL}}$ and retain the ImageNet domain knowledge. As parameters from group $\mathrm{conv5}$ to $\mathrm{conv1}$ are gradually far from the $\mathcal{L}_{\mathrm{KL}}$ supervision, the corresponding selected actions also increase.

On the other hand, to perform dense prediction in semantic segmentation, the extracted feature maps are first forwarded to $\mathrm{conv6\&7}$ and then to $\mathrm{projection}\_\mathrm{upsampling}$. In addition, similar trend holds for the classifier part: as $\mathrm{projection}\_\mathrm{upsampling}$ is the closest group to $\mathcal{L}_{\mathrm{XE}}$, it is assigned with the highest scale factor for learning rate.

\begin{figure*}[!ht]
	\centering
	\resizebox{0.98\textwidth}{!}{
	\begin{tabular}{@{}cccccccccc@{}}
		\cellcolor{city_color_1}\textcolor{white}{~~road~~} &
		\cellcolor{city_color_2}~~sidewalk~~&
		\cellcolor{city_color_3}\textcolor{white}{~~building~~} &
		\cellcolor{city_color_4}\textcolor{white}{~~wall~~} &
		\cellcolor{city_color_5}~~fence~~ &
		\cellcolor{city_color_6}~~pole~~ &
		\cellcolor{city_color_7}~~traffic lgt~~ &
		\cellcolor{city_color_8}~~traffic sgn~~ &
		\cellcolor{city_color_9}~~vegetation~~ & 
		\cellcolor{city_color_0}\textcolor{white}{~~ignored~~}\\
		\cellcolor{city_color_10}~~terrain~~ &
		\cellcolor{city_color_11}~~sky~~ &
		\cellcolor{city_color_12}\textcolor{white}{~~person~~} &
		\cellcolor{city_color_13}\textcolor{white}{~~rider~~} &
		\cellcolor{city_color_14}\textcolor{white}{~~car~~} &
		\cellcolor{city_color_15}\textcolor{white}{~~truck~~} &
		\cellcolor{city_color_16}\textcolor{white}{~~bus~~} &
		\cellcolor{city_color_17}\textcolor{white}{~~train~~} &
		\cellcolor{city_color_18}\textcolor{white}{~~motorcycle~~} &
		\cellcolor{city_color_19}\textcolor{white}{~~bike~~}
	\vspace{0.5mm}
	\end{tabular}
	}
	\includegraphics[width=0.245\textwidth]{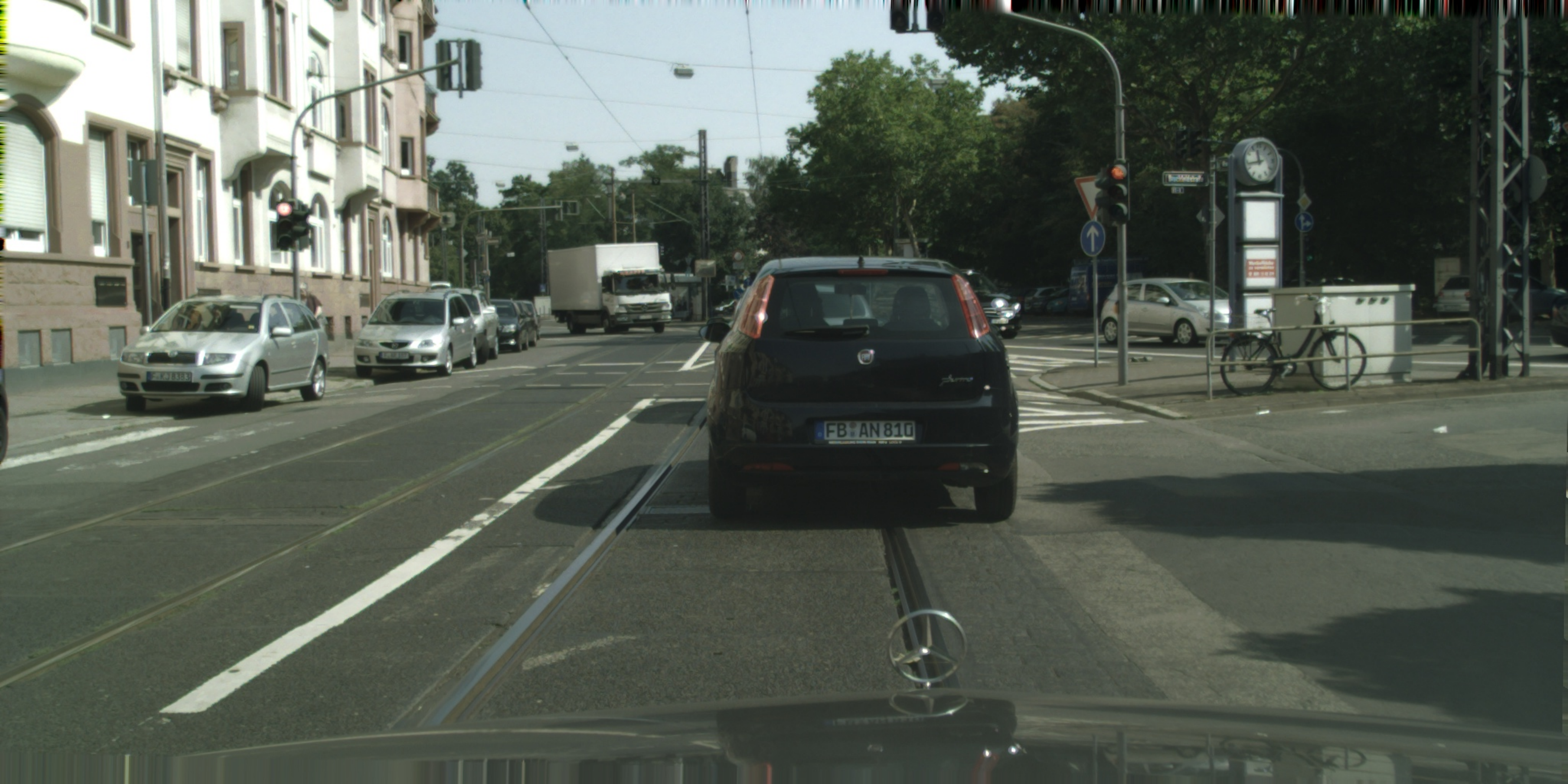}
	\includegraphics[width=0.245\textwidth]{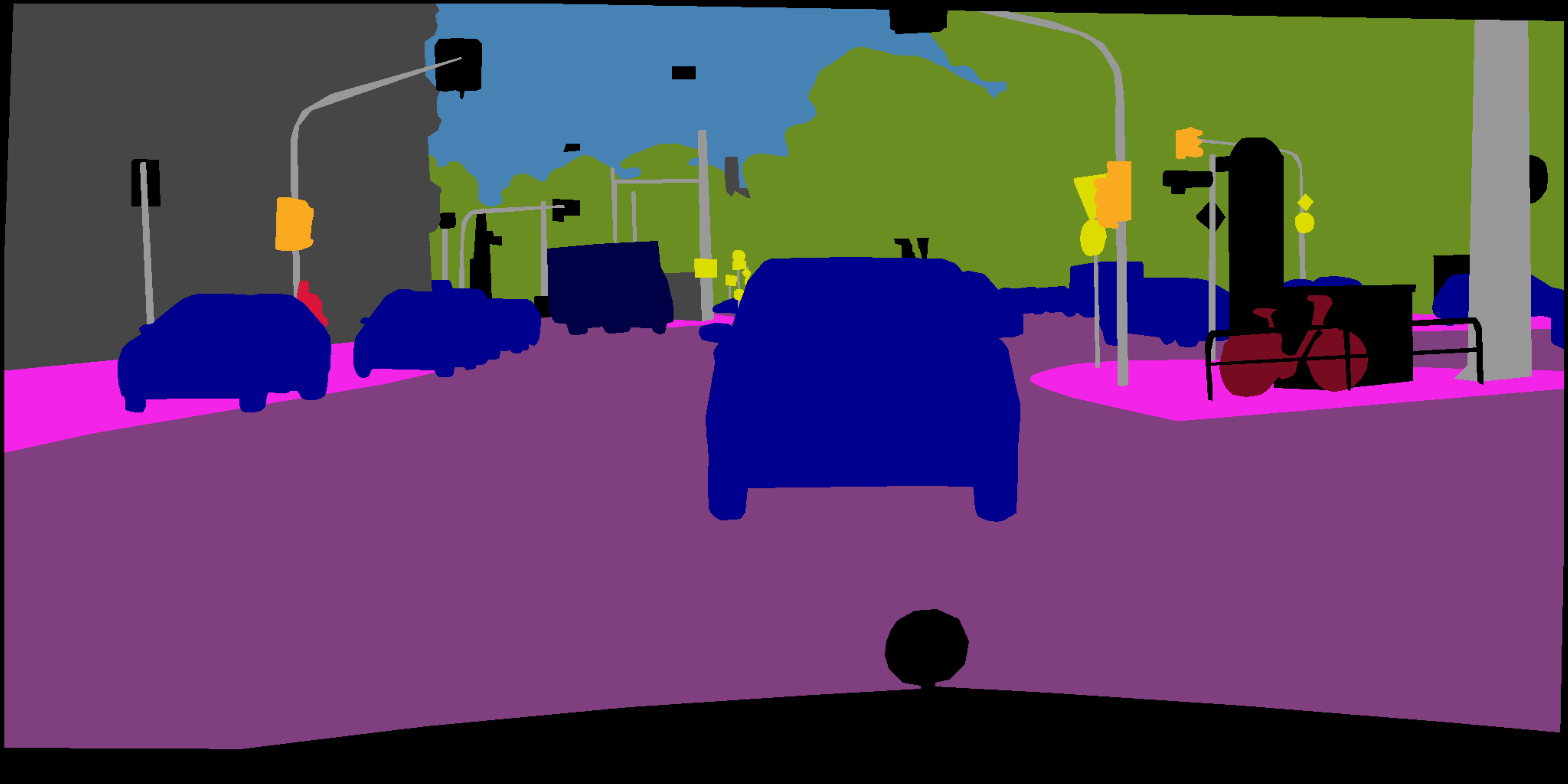}
	\includegraphics[width=0.245\textwidth]{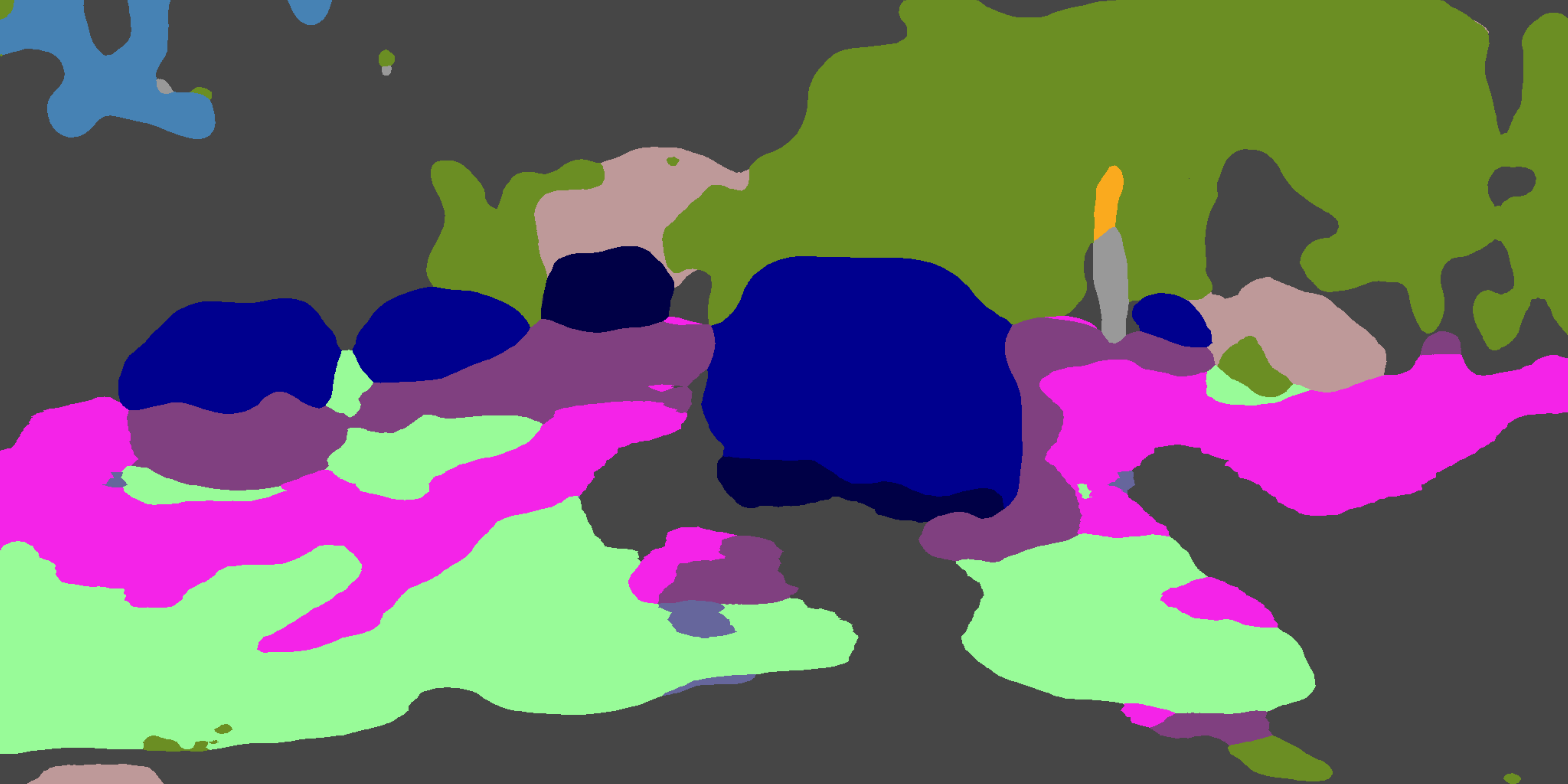}
	\includegraphics[width=0.245\textwidth]{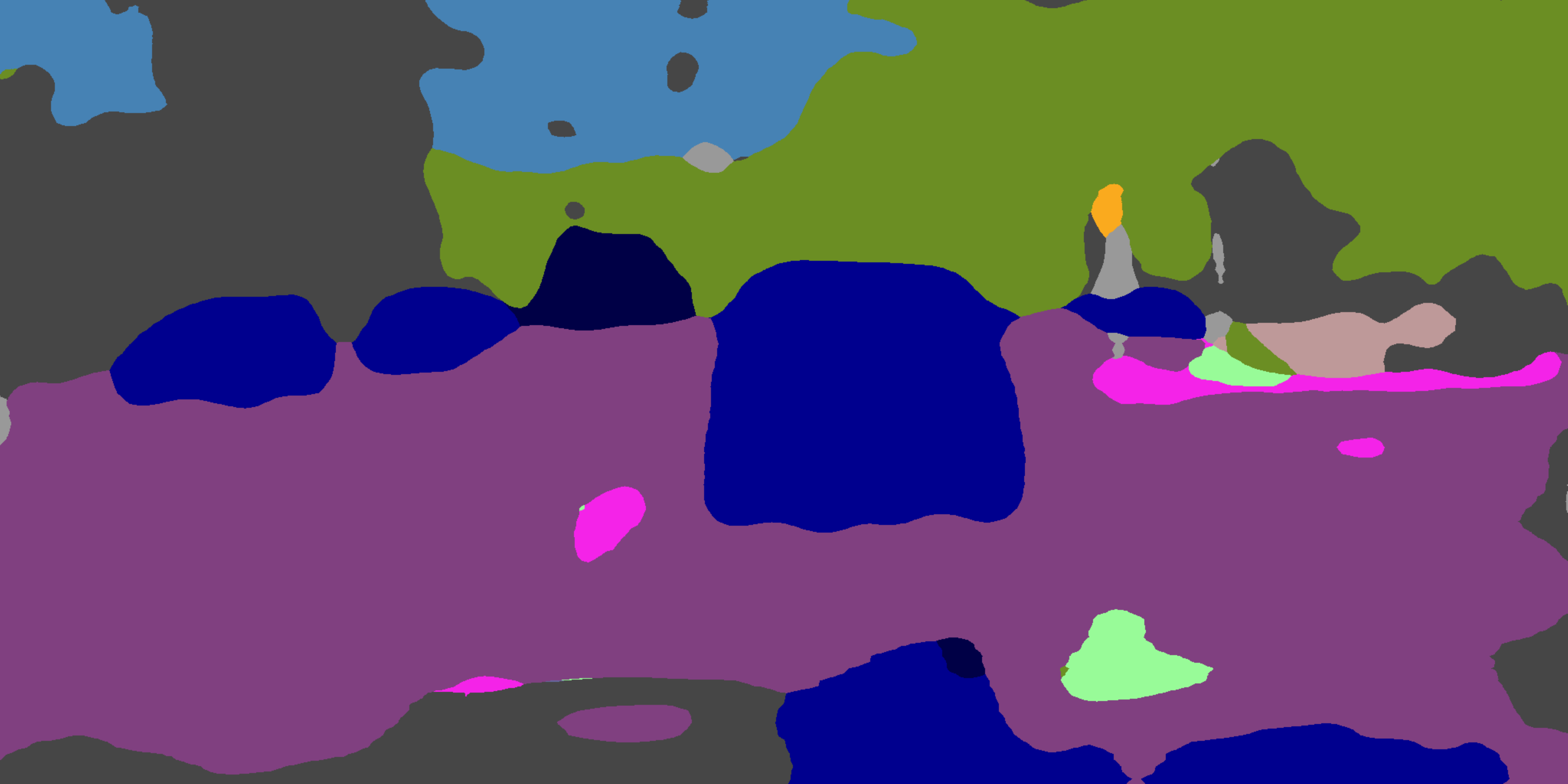}\\\vspace{0.5mm}
	\includegraphics[width=0.245\textwidth]{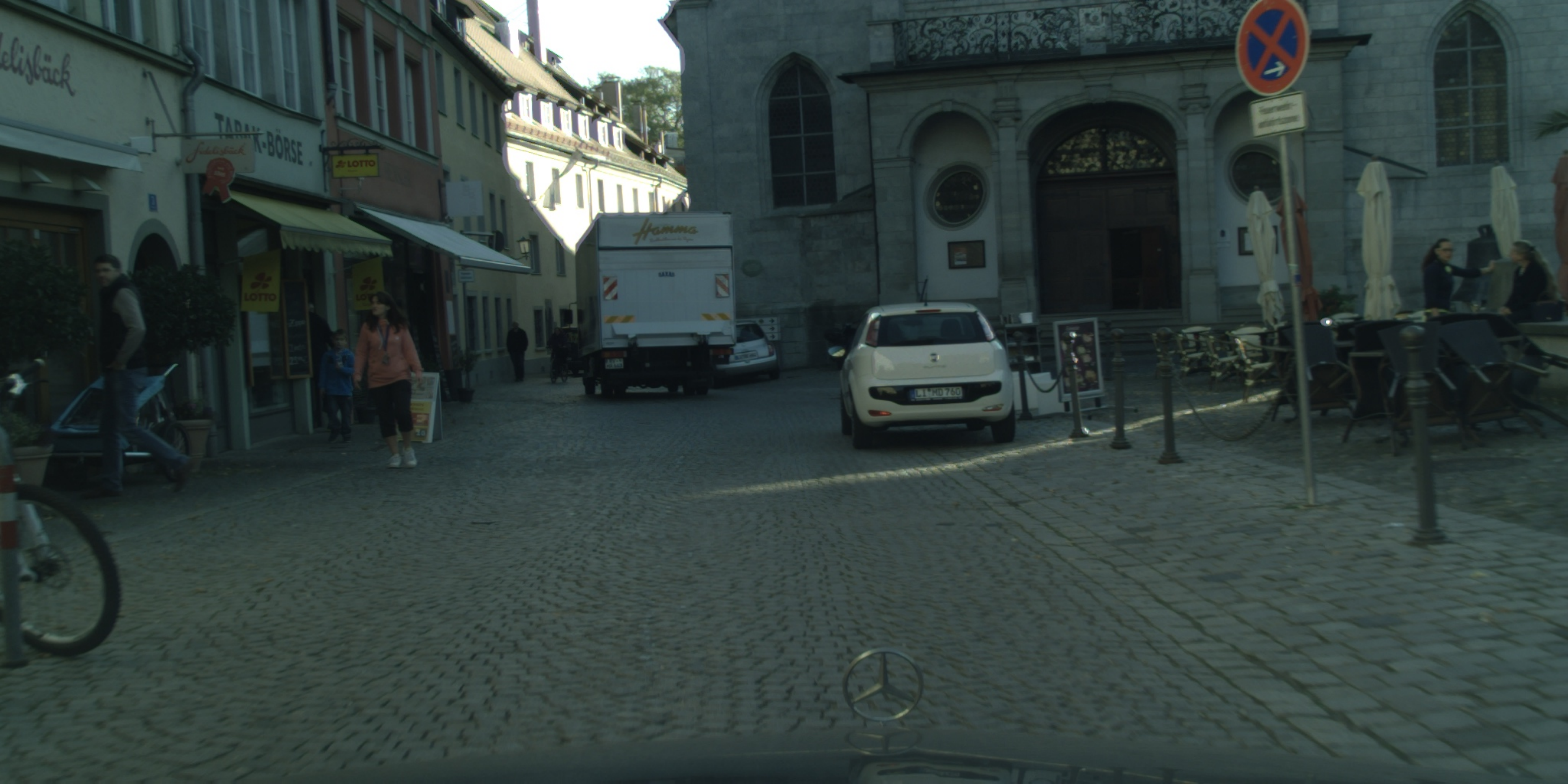}
	\includegraphics[width=0.245\textwidth]{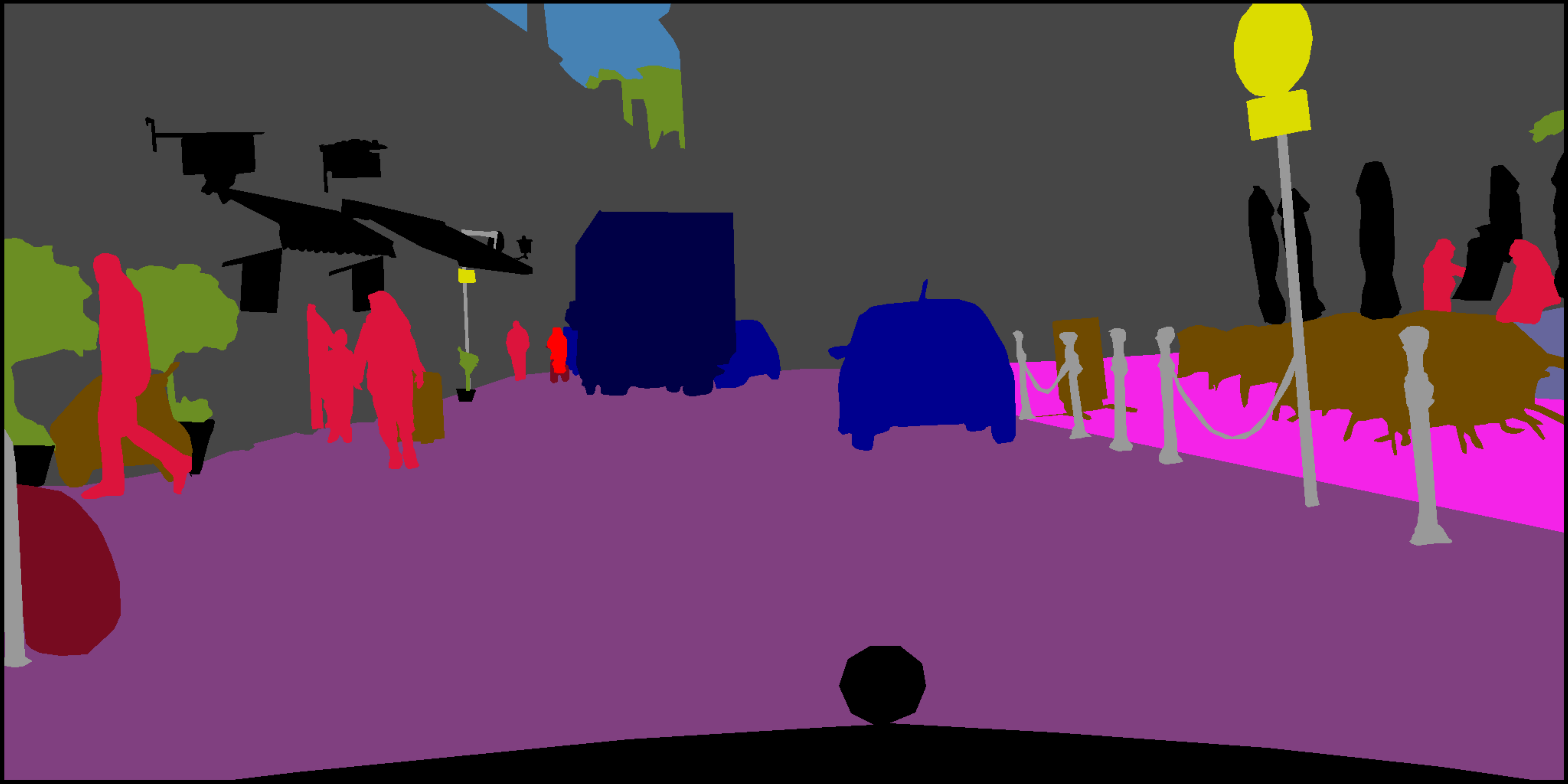}
	\includegraphics[width=0.245\textwidth]{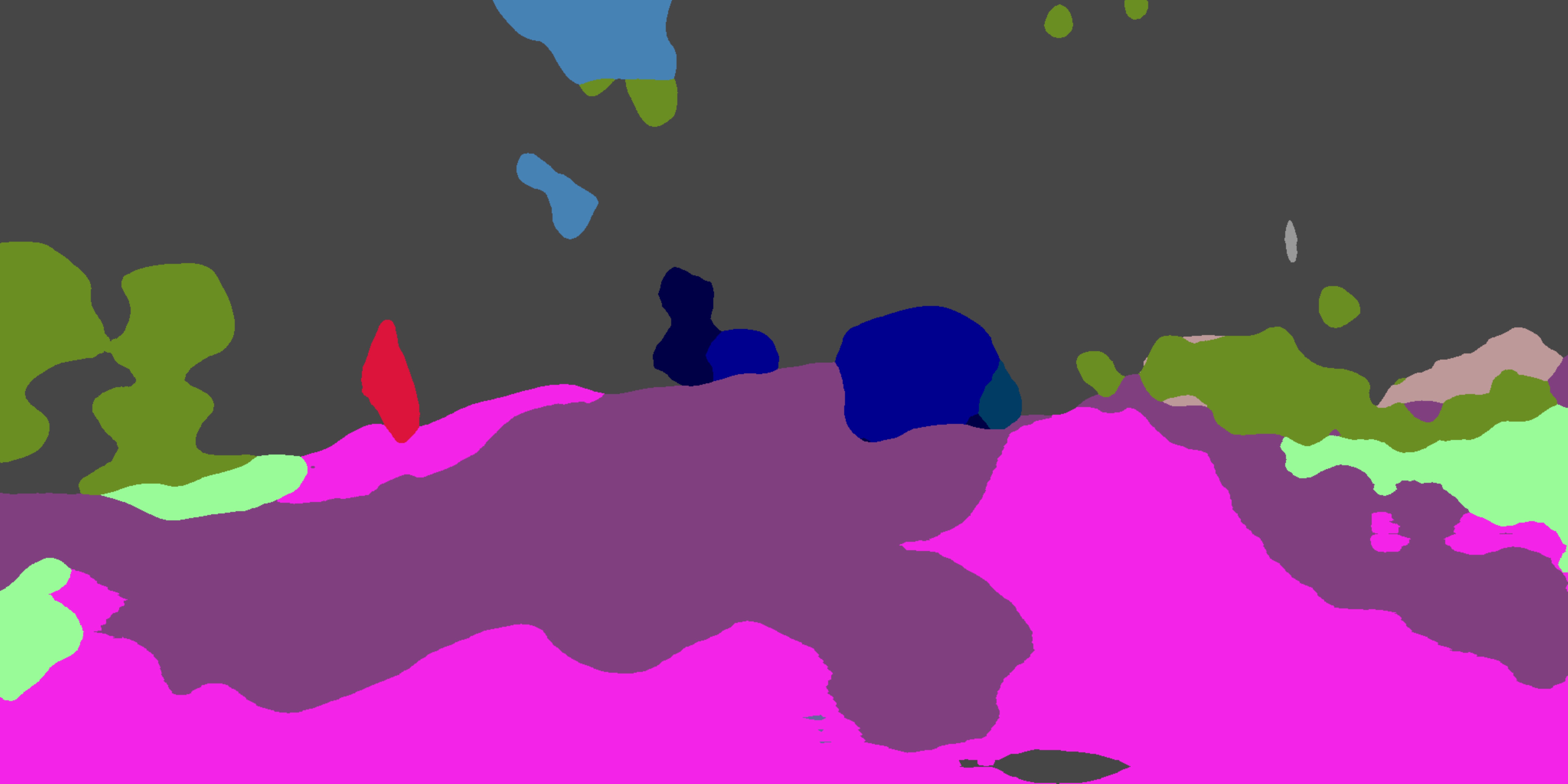}
	\includegraphics[width=0.245\textwidth]{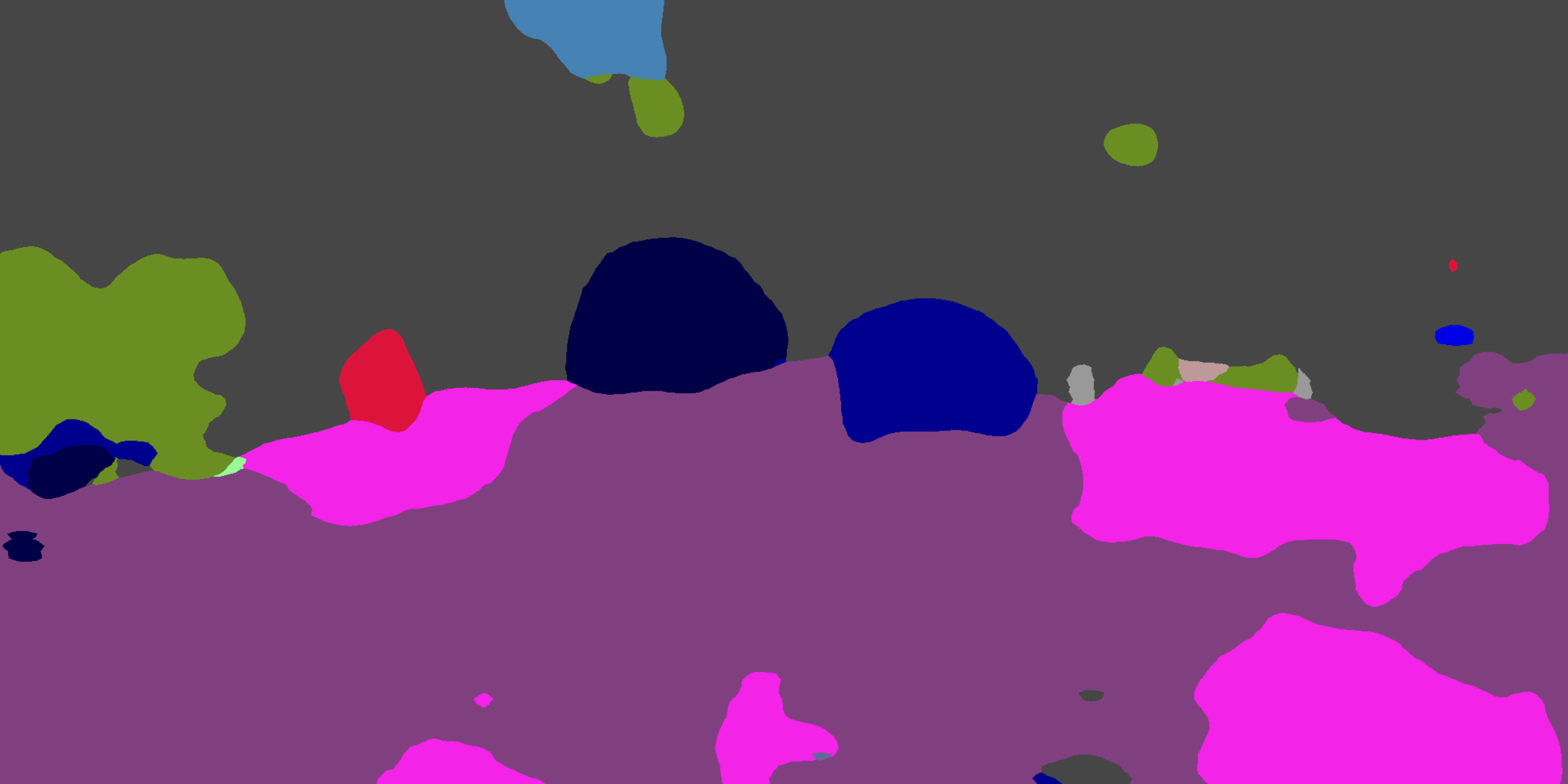}\\\vspace{0.5mm}
	\includegraphics[width=0.245\textwidth]{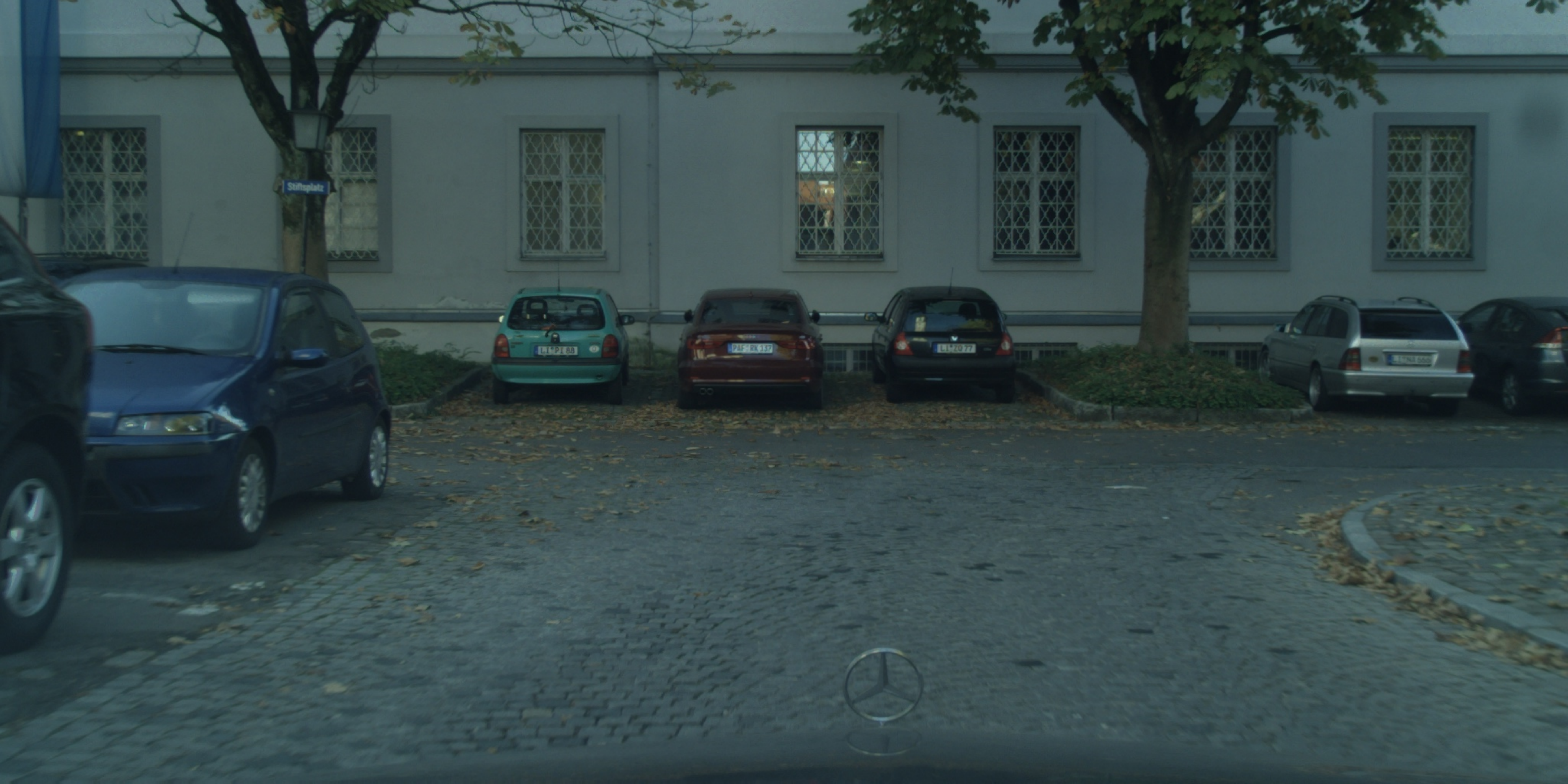}
	\includegraphics[width=0.245\textwidth]{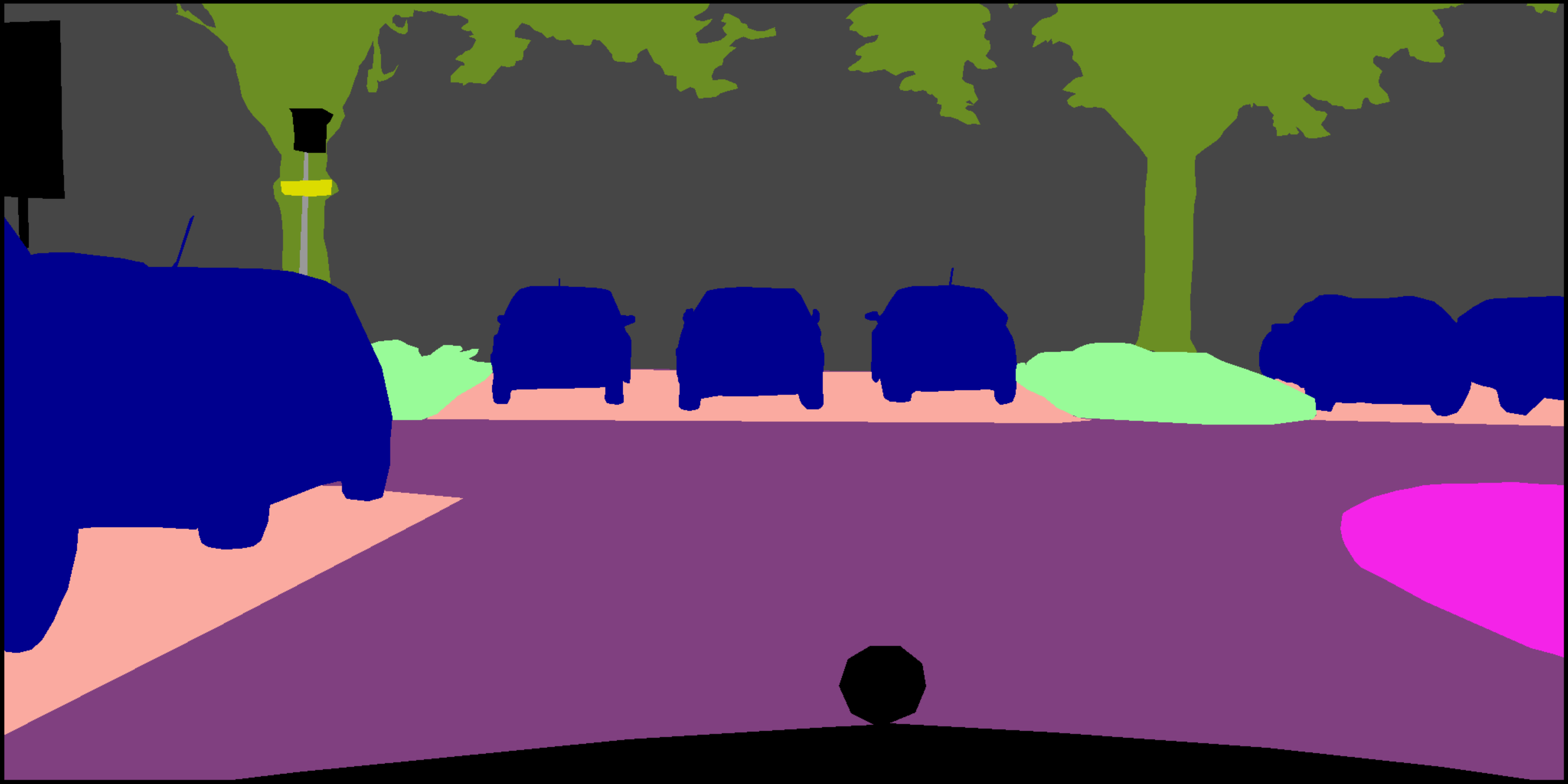}
	\includegraphics[width=0.245\textwidth]{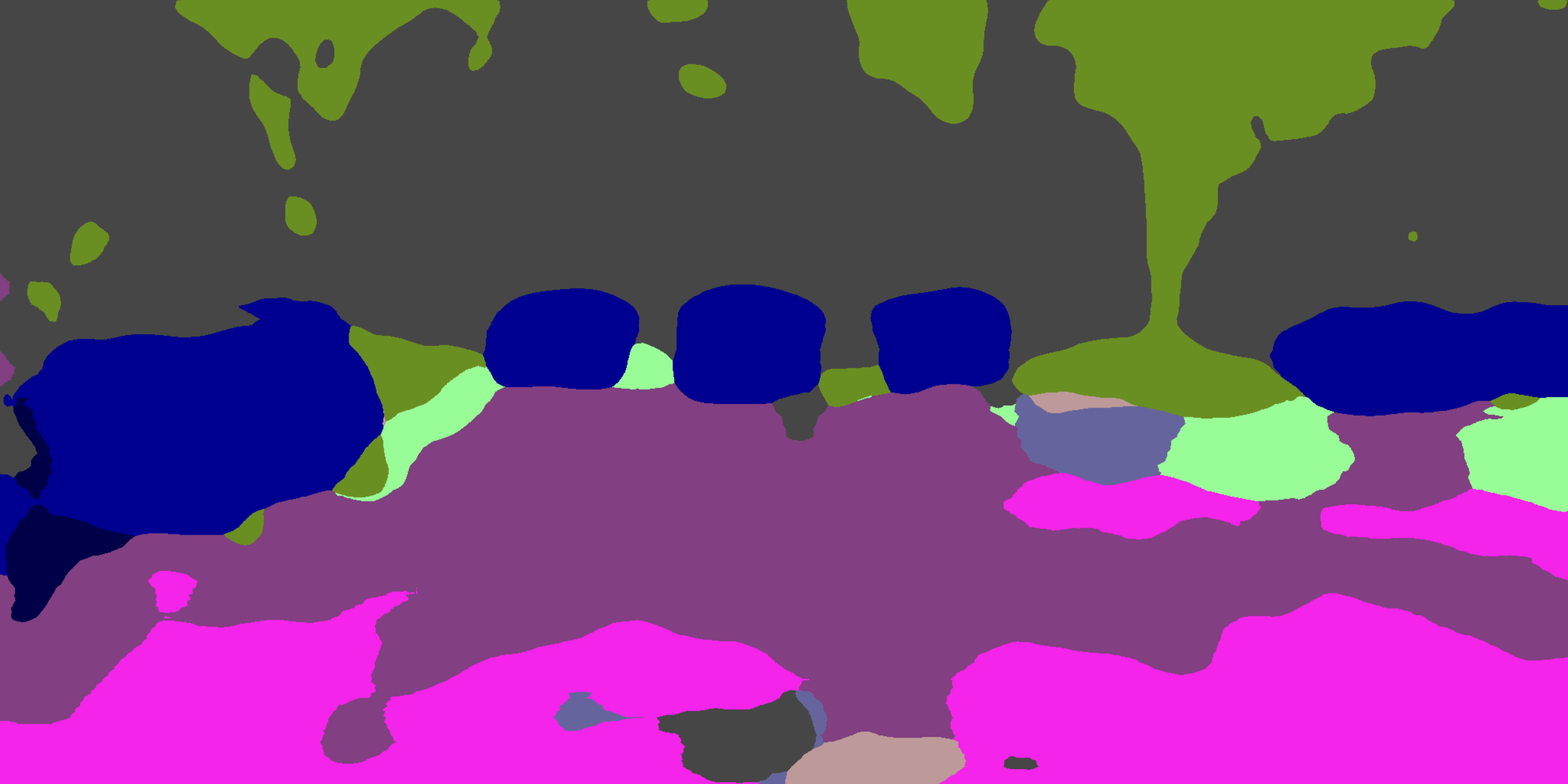}
	\includegraphics[width=0.245\textwidth]{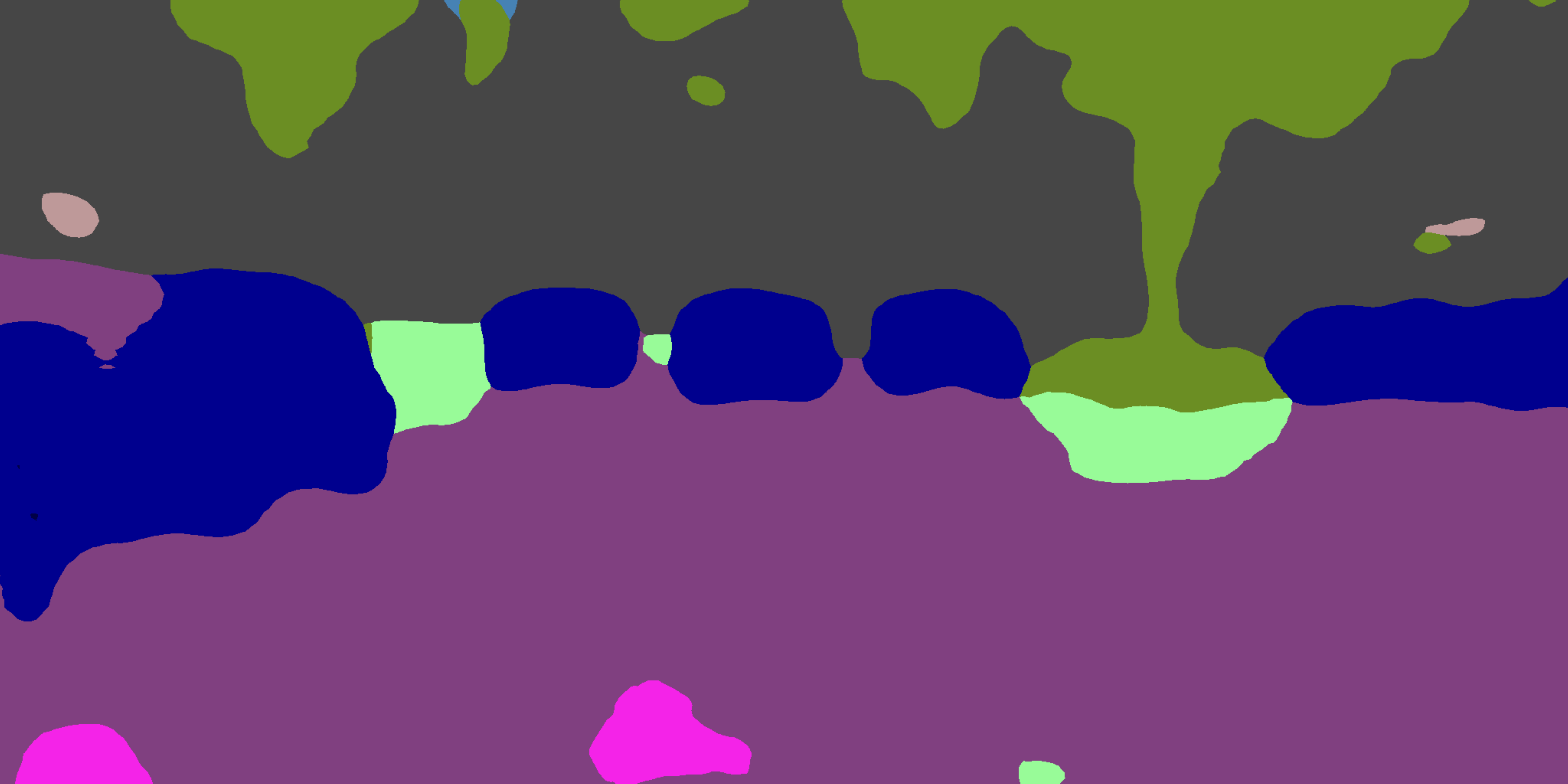}\\
	\caption{Generalization results on GTA5 $\rightarrow$ Cityscapes. Rows correspond to sample images in Cityscapes. From left to right, columns correspond to original images, ground truth, predication results of baseline (FCN-VGG16 \cite{long2015fully}), and prediction by model trained with our ASG framework.}\label{fig:gta2city}
\end{figure*}

\subsection{ASG for Unsupervised Domain Adaptation}

The proposed ASG framework not only can improve the synthetic-to-real generalization performance, but also can considerably benefit downstream tasks such as unsupervised domain adaptation. Here we present synthetic-to-real domain adaptation results on VisDA-17 \cite{peng2017visda} in Table \ref{table:visda_adaptation}, where the model trained by ASG (which did not use any real target images during training) is leveraged as the source model (i.e., starting point for the unsupervised domain adaptation training), and the CBST/CRST frameworks are adopted exactly following \cite{zou2018unsupervised,zou2019confidence} for fair comparison purposes.

Starting from a much better initialization (our 61.1\% compared with 51.6\% in \cite{zou2019confidence}), we significantly boost the adaptation performance over 6\% compared with CBST/CRST, achieving 84.6\% on Visda-17. It is important to emphasize that such improvement is obtained without any extra supervision and external knowledge. The only difference lies in smarter synthetic-to-real source training which ultimately leads to improved adaptation.

\begin{table}[h!]
\vspace{-1em}
\caption{Synthetic-to-real adaptation on Visda-17. We follow the same settings in \cite{zou2019confidence} to set the weights as 0.1 and 0.25 for MRKLD and LRENT respectively, and report the averages and standard deviations (in brackets) of the evaluation results over five runs. Model: ResNet-101. ``Tgt Img'': whether the method leveraged target real images during training. Top-1 accuracies are in percentage (\%).}\vspace{0.3em}
\centering
{\footnotesize
\begin{tabular}{ccc}
\toprule
Method & Tgt Img & Accuracy \\ \midrule
Source \cite{saito2017adversarial} & \singlexmark & 52.4 \\
DANN \cite{ganin2016domain} & \singlecmark & 57.4 \\
MCD \cite{saito2018maximum} & \singlecmark & 71.9 \\
ADR \cite{saito2017adversarial} & \singlecmark & 74.8 \\
SimNet-Res152 \cite{pinheiro2018unsupervised} & \singlecmark & 72.9 \\
GTA-Res152 \yrcite{sankaranarayanan2018generate} & \singlecmark & 77.1 \\ \midrule
Source-Res101 \cite{zou2019confidence} & \singlexmark & 51.6 \\
CBST \cite{zou2018unsupervised} & \singlecmark & 76.4 (0.9) \\
CRST (MRKLD) \yrcite{zou2019confidence} & \singlecmark & 77.9 (0.5) \\
CRST (MRKLD + LRENT) \yrcite{zou2019confidence} & \singlecmark & 78.1 (0.2) \\ \midrule
Source-Res101 (ASG) & \singlexmark & 61.1 \\
ASG + CBST & \singlecmark & 82.5 (0.7) \\
ASG + CRST (MRKLD) & \singlecmark & \textbf{84.6} (0.4) \\
ASG + CRST (MRKLD + LRENT) & \singlecmark & 84.5 (0.4)\\ \bottomrule
\end{tabular}
}\label{table:visda_adaptation}
\end{table}

\begin{figure*}[ht!]
\centering
\resizebox{0.98\textwidth}{!}{
\begin{tabular}{@{}cccccccccccc@{}}
\cellcolor{visda_color_1}{~~aero~~} &
\cellcolor{visda_color_2}~~bike~~&
\cellcolor{visda_color_3}{~~bus~~} &
\cellcolor{visda_color_4}{~~car~~} &
\cellcolor{visda_color_5}~~horse~~ &
\cellcolor{visda_color_6}~~knife~~ &
\cellcolor{visda_color_7}\textcolor{white}{~~motor~~} &
\cellcolor{visda_color_8}\textcolor{white}{~~person~~} &
\cellcolor{visda_color_9}\textcolor{white}{~~plant~~} & 
\cellcolor{visda_color_0}\textcolor{white}{~~board~~}
\cellcolor{visda_color_10}~~train~~ &
\cellcolor{visda_color_11}~~truck~~
\end{tabular}
}
\vspace{1mm}
\includegraphics[width=0.25\textwidth]{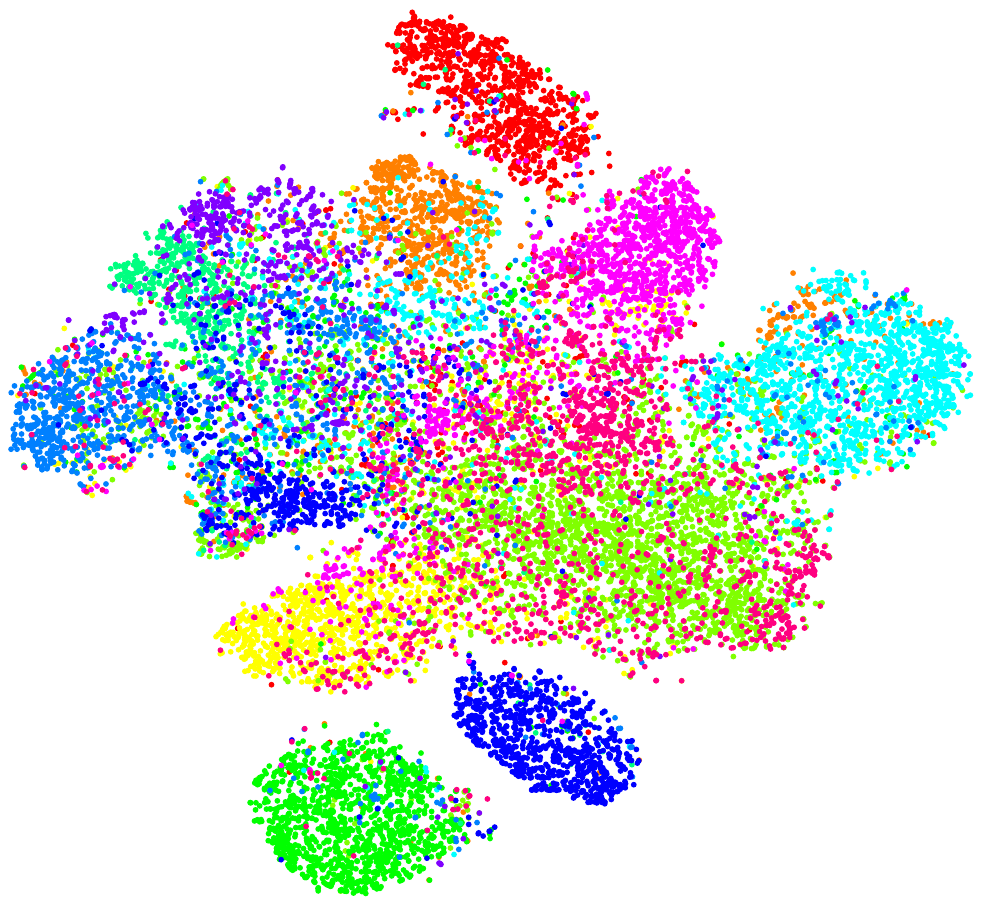}
\includegraphics[width=0.25\textwidth]{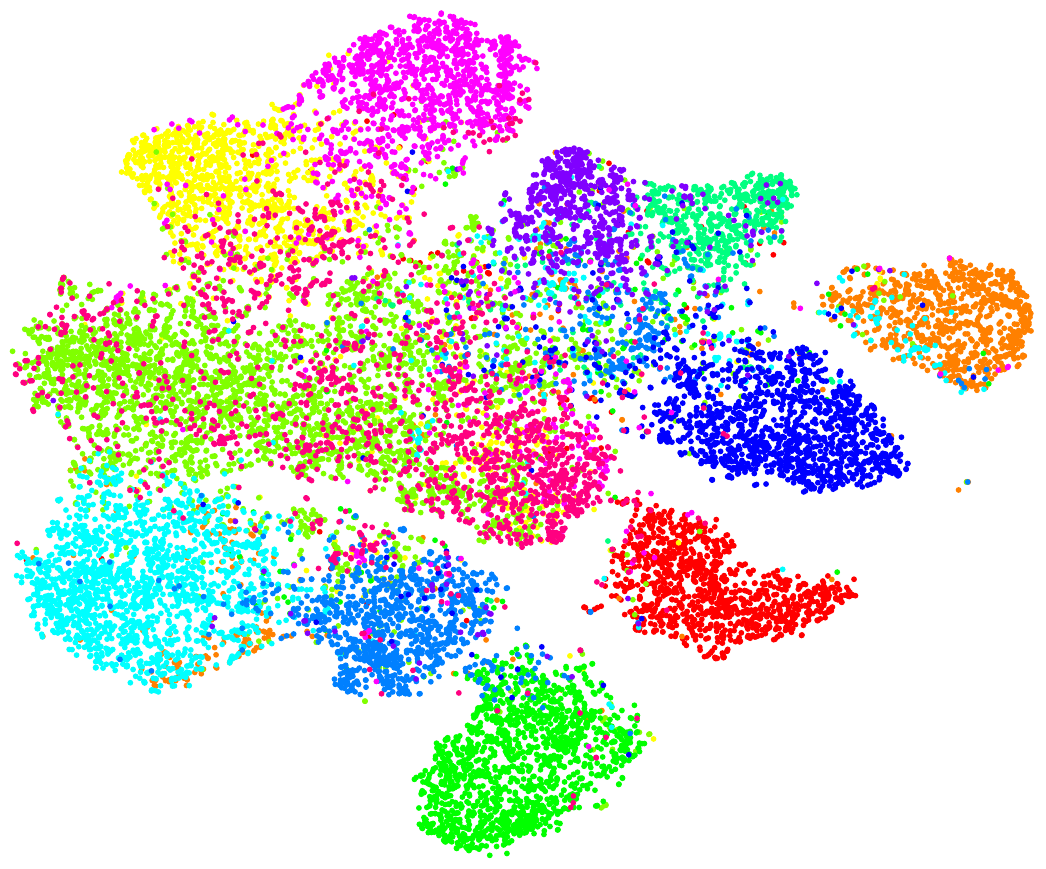}
\includegraphics[width=0.25\textwidth]{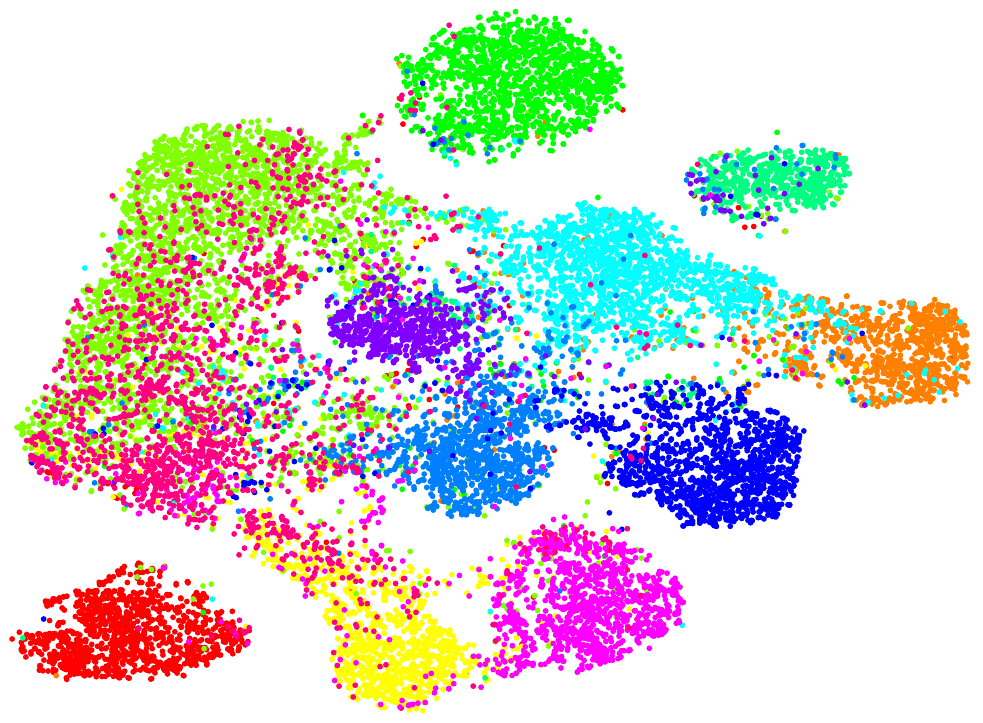}
\includegraphics[width=0.21\textwidth]{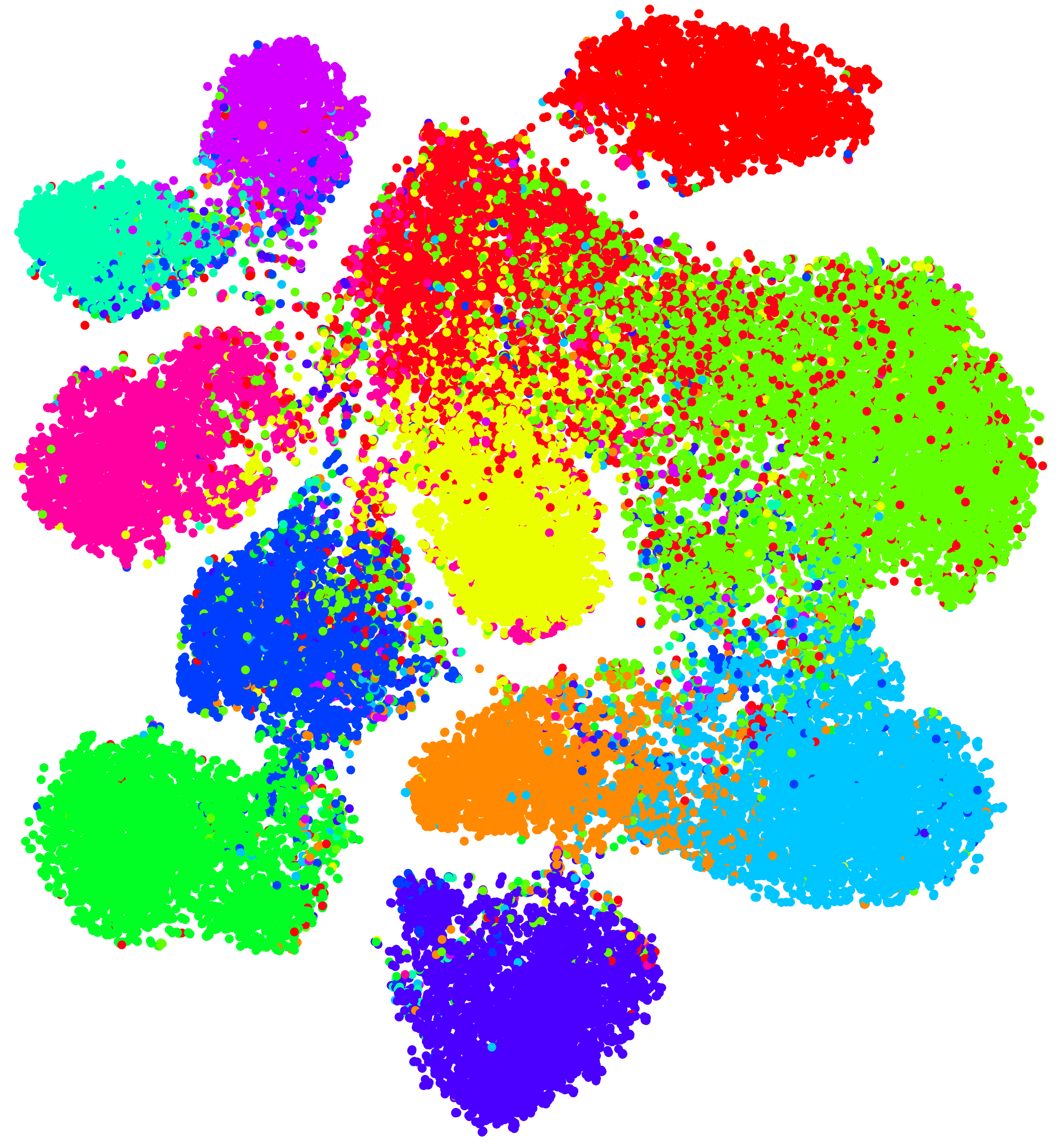}
\caption{t-SNE visualization of feature embeddings of different models on the target domain of VisDA-17. From left to right: source model \cite{zou2019confidence}, CBST \cite{zou2018unsupervised}, CRST (MRKLD+LRENT) \cite{zou2019confidence}, and ASG + CRST (MRKLD+LRENT).}\label{fig:feat_vis}
\end{figure*}

\textbf{Feature visualization.} We show the t-SNE visualization of the feature embeddings extracted by the backbone (ResNet-101) of different models in Fig. \ref{fig:feat_vis}. Compared with Both CBST \cite{zou2018unsupervised} and CRST (MRKLD+LRENT) \cite{zou2019confidence}, feature embeddings obtained by ASG + CRST form purer clusters in terms of semantic labels.

\section{Related Work}

\subsection{Domain Generalization and Adaptation}
Domain generalization considers the problem of generalizing a model on the unseen target domain without leveraging any target domain images \cite{gan2016learning,muandet2013domain,yuan2020calibrated}.
Muandet et al. \yrcite{muandet2013domain} proposed to use the MMD (Maximum Mean Discrepancy) to align the distributions from different domains and train the network with adversarial learning. Li et al. \yrcite{li2017deeper} built separate networks for each source domain and used the shared parameters for the test. Li et al. \yrcite{li2018learning} improved the generalization performance by using a meta-learning approach on the split training sets. Pan et al. \yrcite{pan2018two} boosted a CNN’s generalization by carefully integrating the Instance Normalization and Batch Normalization as building blocks.

Unsupervised domain adaptation (UDA) trains a model towards a specific target domain, where the (unlabeled) images from the target domain are available for training. One major idea is to learn domain invariant embeddings by minimizing the distribution divergence between the source and target domain \cite{long2015learning,sun2016deep,tzeng2014deep}. Hoffman et al. \yrcite{hoffman2017cycada} reduced domain gap by first translating the source images into target style with a cycle consistency loss, and then aligning the feature maps of the network across different domains through the adversarial training. Other works that leverage image level translation to bridge the domain gap include domain stylization~\cite{dundar2020domain} and DLOW~\cite{gong2019dlow}. Besides image-level translation, a number of works also perform adversarial learning at feature~\cite{saito2018adversarial,chen2019learning,liu2019feature} or output level~\cite{Tsai_adaptseg_2018} for the improved domain adaptation performance. In addition, Zou et al. \yrcite{zou2018unsupervised,zou2019confidence} proposed an expectation-maximization like UDA framework based on an iterative self-training process, where the loss of the latent variable is minimized. This is achieved by alternatively generating pseudo labels on target data and re-training the model with the mixed source and pseudo target labels.

In contrast to the above existing domain generalization and adaptation methods, we resort to leveraging the ImageNet pre-trained model as a proxy guidance during the synthetic-to-real transfer learning, without any extra adversarial training or modification to model architecture.

\subsection{Lifelong Learning}

Lifelong learning \cite{thrun1998lifelong} focuses on flexibly appending new tasks to the model's training schedules, while maintaining the knowledge captured from previous old tasks. Li \& Hoiem \yrcite{li2017learning} leverages only new task data to train the network while preserving the original capabilities by minimizing the outputs between the old network and the newly learned one. Lopez-Paz and Ranzato \yrcite{lopez2017gradient} proposed a Gradient Episodic Memory (GEM) to alleviate the knowledge forgetting while transferring knowledge from previous tasks. Shin et al. \yrcite{shin2017continual} developed a Deep Generative Replay framework, which is used to sample training data from previous tasks when training the new task. A number of other works on lifelong learning with related or similar applications include \cite{zenke2017continual,kirkpatrick2017overcoming,shafahi2019adversarially} where lifelong learning is shown to avoid catastrophic forgetting and benefit tasks such as incremental tasks learning, domain adaptation and adversarial defense. One work that is particularly related to our synthetic-to-real generalization theme is \cite{chen2018road} where the authors propose a spatial aware adaptation scheme and also leverage a distillation loss to avoid overfitting to synthetic data. Our work differs from the above prior works by carefully looking into the important role played by layer-wise learning rate policies in synthetic-to-real transfer learning problems and accordingly propose a principled solution to automate the policy search.

\subsection{Learning to Optimize}

Andrychowicz et al. \yrcite{andrychowicz2016learning} proposed the first learning-to-optimize framework, where both the optimizee's gradients and loss function values were formulated as the input features for a Recurrent neural network (RNN) optimizer. Their RNN optimizer adopted coordinate-wise weight sharing to alleviate the dimensionality challenge. Li and Malik \yrcite{li2016learning} used the gradient history and objective values as observations and step vectors as actions in their reinforcement learning framework. Chen et al. \yrcite{chen2017learning} leveraged RNN to train a meta-optimizer to optimize black-box functions (e.g. Gaussian process bandits). Recently, Wichrowska et al. \yrcite{wichrowska2017learned} introduced an optimizer of multi-level hierarchical RNN architecture augmented with additional architectural features, in order to improve the generalizability of the optimization tasks. \cite{cao2019learning,you20202} further extended learned optimizers to handling Bayesian swarm optimization, and graph network training, respectively. In our work, we leverage the learning-to-optimize approach to control the layer-wise learning rates for the training of deep CNNs, where the deep CNN (i.e. optimizee) will be transferred from the synthetic source domain to the real target domain, extending the application range of the current learning-to-optimize methods.

\section{Conclusion}

In this paper, we present an Automated Synthetic Generalization (ASG) method for the synthetic-to-real transfer learning problem. We carefully analyzed the pitfall in existing generalization approaches where the ImageNet domain knowledge is catastrophically forgotten. By leveraging the minimization of predictions between ImageNet pre-trained model and the model for the new task as a proxy guidance, the generalization performance is dramatically improved during the whole training process. We further include a reinforcement learning based learning-to-optimize strategy to automate the layer-wise learning rates towards a better generalization performance. Our experiments demonstrate both the superior generalization performance and the automated learning schedules by our ASG framework.

\section{Acknowledge}
Work done during internship at NVIDIA. We appreciate the computing power supported by NVIDIA GPU infrastructure. We also thank for the discussion and suggestions from four anonymous reviewers and the help from Yang Zou for the domain adaptation experiments. The research of Z. Wang was partially supported by NSF Award RI-1755701.

\bibliography{ref}
\bibliographystyle{icml2020}

\end{document}